\documentclass{article}

\usepackage{arxiv}

\usepackage[utf8]{inputenc}
\usepackage[T1]{fontenc}
\usepackage{amsmath,amssymb}
\usepackage{graphicx}
\usepackage{cite}
\usepackage{booktabs}
\usepackage{multirow}
\usepackage{tabularx}
\usepackage{array}
\usepackage[table]{xcolor}
\usepackage[normalem]{ulem}
\usepackage[caption=false,font=footnotesize]{subfig}
\usepackage{url}
\usepackage[hidelinks]{hyperref}

\definecolor{toolbg}{RGB}{244,248,252}
\definecolor{oraclebg}{RGB}{250,245,238}
\usepackage{pifont}
\newcommand{\cmark}{\textcolor{green!60!black}{\ding{51}}}
\newcommand{\xmark}{\textcolor{red!70!black}{\ding{55}}}
\usepackage{authblk}

\renewcommand{\shorttitle}{\textit{arXiv}: Vertical Structure}

\begin{document}

\title{Beyond Appearance: Can Multimodal Large Language Models Exploit Vertical Structure for Remote Sensing Natural Scene Understanding?\thanks{Jing Huang and Duanchu Wang contributed equally to this work.}}
\date{}

\author[1]{Jing Huang\textsuperscript{*}}
\author[1]{Duanchu Wang\textsuperscript{*}}
\author[2]{Junjie Yang}
\author[1]{Zihang Cheng}
\author[1]{Cheng Li}
\author[1]{Lin Cui}
\author[1]{Zhouyi Wu}
\author[1,3]{Di Wang\thanks{Corresponding author: Di Wang (e-mail: diwang@xjtu.edu.cn).}}
\affil[1]{School of Software Engineering, Xi'an Jiaotong University, Xi'an 710049, China}
\affil[2]{Shenyang Institute of Automation, University of Chinese Academy of Sciences, Shenyang, China}
\affil[3]{State Key Laboratory of Human-Machine Hybrid Augmented Intelligence, Xi'an Jiaotong University}

\maketitle

\begin{abstract}
Multimodal large language models (MLLMs) have advanced rapidly in remote-sensing analysis, yet existing evaluations remain predominantly 2D-centric. Because spectrally confused regions can appear nearly identical yet differ substantially in vertical structure, appearance alone is often insufficient for reliable semantic interpretation in natural scenes. Vertical structure therefore provides decision-critical physical evidence, yet whether current MLLMs can effectively perceive, ground, and utilize such geometric evidence remains underexplored. 
To bridge this gap, we introduce VertiCue-Bench, the first diagnostic benchmark that uses controlled interventions to probe whether vertical height evidence is actually perceived, grounded, and utilized, and we establish a three-stage evidence-utilization framework of Perception--Grounding--Utilization. By constructing a Representation Intervention Spectrum spanning multiple presentation and interaction modalities, including Raw Visual, Tool-assisted, and Oracle Text conditions, together with controlled counterfactual tests, we conduct an in-depth disentangled diagnosis across 10 state-of-the-art models. Our experiments reveal and formally characterize the Vertical Structure Utilization Gap. Although current models exhibit emerging geometric perception capabilities, they still struggle to accurately ground vertical evidence to relevant spatial entities and integrate it into high-level semantic decisions. This finding identifies a critical bottleneck in developing physically grounded and geometry-aware remote-sensing MLLMs.
\end{abstract}


\section{Introduction}
Multimodal large language models (MLLMs) have rapidly extended remote-sensing analysis from image recognition and captioning to visual question answering, visual grounding, and multi-step geospatial reasoning~\cite{kuckreja2023geochat,muhtar2024lhrsbot,bazi2024rsllava,pang2025vhm,shabbir2025geopixel,liu2026rsthinker,yao2026remotereasoner}. Recent benchmarks have accordingly expanded toward increasingly diverse scenes and reasoning tasks~\cite{hu2025rsgpt,li2024hrvqa,wang2024earthvqa,li2024vrsbench,danish2025geobench,wang2025xlrsbench}. Nevertheless, most existing evaluations remain fundamentally 2D-centric, implicitly assuming that decision-relevant evidence can be recovered from optical appearance. This assumption breaks down in natural scenes, where spectral confusion can make ecologically distinct regions appear highly similar in color, spectral response, and texture~\cite{lu2007survey}. Low shrubs, regenerating woodland, and mature forest, for example, may be almost indistinguishable in 2D imagery while differing markedly in canopy height and vertical organization. In such cases, optical appearance alone can mislead an interpretation of the underlying ecological or structural category.

In such cases, vertical structure becomes decision-critical physical evidence rather than merely an auxiliary cue. Height-sensitive measurements and derived products such as canopy height models (CHMs) explicitly characterize canopy height and structural variation that may not be reliably recovered from red-green-blue (RGB) imagery alone~\cite{Balestra2024,Wu2024,Ma2024}. In operational remote-sensing analysis, such information provides a physical basis for distinguishing vegetation maturity, vertical contrast, shelter conditions, and other height-constrained properties that remain ambiguous in 2D appearance. Despite its importance, whether current MLLMs can transform explicitly provided height measurements into reliable evidence for subsequent semantic reasoning remains insufficiently studied.

This motivates the central research question. \emph{When 2D texture appearance alone is insufficient to support a decision, can current MLLMs effectively exploit vertical geometric evidence for reliable reasoning and decision-making?} 
Addressing this question requires examining not only whether height information can be read, but whether it can ultimately support a reliable decision. Consider, for example, a candidate region that visually resembles forest but whose measured mean canopy height violates a required height range.
Solving such a question requires more than recognizing the semantic category ``forest''; the model must associate the measured height with the same candidate region and enforce the geometric condition as a decision constraint. Therefore, the central research question naturally decomposes into three evaluation questions.
First, can an MLLM reliably acquire task-relevant physical quantities from a height representation?
Second, can it associate those quantities with the correct visual entities?
Third, after the physical evidence is made explicit and correctly associated, can the model allow it to alter a semantic decision when it conflicts with appearance-based priors?

To answer these three questions, we formulate vertical-structure utilization as a three-stage evidence-utilization framework. Perception recovers the physical quantity from the representation, Grounding binds it to the corresponding visual entity, and Utilization converts the grounded quantity into a decision constraint. These stages form a sequential evidence-utilization chain, yet end-to-end accuracy conflates them. Because an incorrect answer may originate at any stage, while a correct answer may reflect appearance priors or interface shortcuts rather than genuine height use, a high score alone cannot establish whether vertical evidence was actually used. We therefore probe the three questions separately and add matched counterfactual perturbations that selectively change or preserve the relevant height condition, giving a more direct behavioral test of evidence dependence.

In this work, we introduce VertiCue-Bench, the first diagnostic benchmark for remote-sensing natural-scene understanding that uses controlled interventions to probe whether vertical height evidence is actually perceived, grounded, and utilized. Built from pixel-aligned National Agriculture Imagery Program (NAIP) RGB imagery and CHMs, it contains 1,534 spatially aligned instances across 17 fine-grained tasks, progressing from height perception and spatial grounding to ambiguity-aware semantic--geometric reasoning. Beyond the canonical passive visual interface, we construct a Representation Intervention Spectrum spanning Raw Visual, Tool-assisted, and Oracle Text conditions that progressively externalize the Perception and Grounding stages. We further add presentation interventions, controlled counterfactual height swaps, and semantic--geometric conflict analysis to test whether model decisions genuinely depend on vertical evidence. As summarized in Table~\ref{tab:dataset_comparison}, VertiCue-Bench distinguishes itself from prior benchmarks by jointly incorporating pixel-aligned continuous height fields as explicit physical evidence, stage-wise capability diagnosis, and counterfactual evaluation.

\begin{table}[t]
\centering
\captionsetup{skip=1pt}
\caption{Comparison of VertiCue-Bench with representative remote-sensing and height-aware MLLM benchmarks.}
\label{tab:dataset_comparison}
\setlength{\tabcolsep}{4.5pt}
\renewcommand{\arraystretch}{1.10}
\resizebox{\textwidth}{!}{%
\begin{tabular}{l c l c c c c c l}
\toprule
\multirow{2}{*}{Dataset} &
\multirow{2}{*}{Test/Eval. Size} &
\multirow{2}{*}{Primary Focus} &
\multicolumn{3}{c}{Vertical Evidence} &
\multicolumn{2}{c}{Diagnostic Design} &
\multirow{2}{*}{Remarks} \\
\cmidrule(lr){4-6}\cmidrule(lr){7-8}
& & &
Height-Aware &
Explicit Height &
Pixel-Aligned Field &
Stage-Wise &
Counterfactual &
\\
\midrule

VRSBench~\cite{li2024vrsbench}
& 37,408 QA
& RS image understanding
& \xmark & \xmark & \xmark
& \xmark & \xmark
& VQA test split \\

GEOBench-VLM~\cite{danish2025geobench}
& $>$10,000
& Geospatial capabilities
& \xmark & \xmark & \xmark
& \xmark & \xmark
& Broad geospatial evaluation \\

CHOICE~\cite{an2025choice}
& 10,507
& Hierarchical RS capabilities
& \xmark & \xmark & \xmark
& \xmark & \xmark
& Perception / reasoning taxonomy \\

XLRS-Bench~\cite{wang2025xlrsbench}
& 32,389 QA
& UHR RS understanding
& \xmark & \xmark & \xmark
& \xmark & \xmark
& VQA evaluation set \\

Geo3DVQA~\cite{tsujimoto2026geo3dvqa}
& 3,000 QA
& RGB-to-3D reasoning
& \cmark & \cmark & \cmark
& \xmark & \xmark
& RGB-only primary setting; DSM ablations \\

SpatialSky-Bench~\cite{zhang2026spatialsky}
& $\sim$1,000 QA
& UAV spatial intelligence
& \cmark & \xmark & \xmark
& \xmark & \xmark
& Height / distance judgments \\

\rowcolor{purple!5}
\textbf{VertiCue-Bench (Ours)}
& \textbf{1,534}
& \textbf{Vertical-structure utilization}
& \textbf{\cmark} & \textbf{\cmark} & \textbf{\cmark}
& \textbf{\cmark} & \textbf{\cmark}
& \textbf{Explicit CHM evidence + controlled diagnosis} \\

\bottomrule
\end{tabular}
}
\end{table}

Extensive evaluation across 10 state-of-the-art general-purpose and remote-sensing-specialized MLLMs reveals a consistent Vertical Structure Utilization Gap. Although current models exhibit emerging capability for low-level geometric perception, this information does not reliably propagate through the full Perception--Grounding--Utilization chain. Models remain particularly fragile when height evidence must be spatially co-registered with visual entities and further incorporated as an effective constraint in high-level semantic decisions. Controlled interventions additionally reveal sensitivity to evidence presentation, incomplete prediction updating under decision-relevant height changes, and persistent preference for 2D appearance when semantic and geometric evidence conflict. These findings suggest that the central bottleneck is not simply whether MLLMs can access height information, but whether they can transform explicit physical measurements into grounded constraints that meaningfully govern their decisions.

The main contributions of this work are the following.
\begin{itemize}
    \item \textbf{Evaluation Target and Benchmark.} We introduce the first diagnostic benchmark for remote-sensing natural-scene understanding that uses controlled interventions to probe whether vertical height evidence is actually perceived, grounded, and utilized. It comprises 1,534 spatially aligned instances and 17 fine-grained tasks that systematically evaluate how MLLMs perceive, ground, and utilize vertical height evidence beyond 2D appearance.

    \item \textbf{Perception--Grounding--Utilization Framework.} We formulate vertical-structure utilization as a three-stage evidence-utilization framework that decomposes the evidence-utilization process into height perception, spatial co-registration, and evidence-driven semantic decision-making, enabling stage-wise diagnosis beyond conventional end-to-end evaluation.

    \item \textbf{Representation Intervention and Utilization Gap.} We construct a step-wise Representation Intervention Spectrum spanning Raw Visual, Tool-assisted, and Oracle Text conditions, together with presentation, controlled counterfactual, and semantic--geometric diagnostics. Experiments across 10 MLLMs formally reveal and characterize the Vertical Structure Utilization Gap, exposing systematic bottlenecks in propagating physical height evidence from low-level perception to grounded high-level decision-making.
\end{itemize}

\section{Related Work}

\subsection{Remote Sensing Multimodal Large Language Models}

Remote sensing vision--language research has progressed from large-scale representation learning and vision--language foundation models~\cite{mall2024graft,liu2024remoteclip,zhang2024rs5m} to instruction-following MLLMs that support captioning, visual question answering, visual grounding, and geospatial reasoning~\cite{kuckreja2023geochat,muhtar2024lhrsbot,bazi2024rsllava,pang2025vhm,soni2025earthdial,wang2025geollava8k,shabbir2025geopixel,liu2026rsthinker,yao2026remotereasoner}. 
Across this progression, however, the evidence modality remains dominated by 2D optical appearance. Conventional remote-sensing studies have shown that LiDAR-derived height and structural measurements can complement optical imagery in forest and vegetation analysis~\cite{Balestra2024,Wu2024,Ma2024}, establishing the value of vertical physical information beyond appearance.
Yet current remote-sensing MLLMs are primarily trained and evaluated to interpret imagery rather than to consume such explicit physical measurements. Existing work therefore primarily answers whether an MLLM understands remote-sensing imagery, whereas our question is whether it can use an explicit physical measurement as decision-relevant evidence.

\subsection{Multimodal Geometric and Physical Reasoning}

Beyond remote sensing, a growing line of work studies geometric and physical reasoning in multimodal models.
Monocular depth estimation provides depth-based geometric cues for downstream perception~\cite{yang2024depthanything}, while spatial and 3D reasoning capabilities have been injected into vision-language models through structured spatial annotations, 3D scene representations, and geometry-aware feature encoding~\cite{chen2024spatialvlm,hong2023dllm}.
These studies demonstrate the value of geometric or physical cues in general-domain reasoning.
They generally do not isolate, however, whether a model can acquire, spatially bind, and decisionally utilize a measured physical quantity as a constraint, which is exactly the question that motivates the controlled diagnostic design of VertiCue-Bench.

\subsection{Remote Sensing 3D and Height-Aware Benchmarks}

Remote sensing benchmarks have expanded from image captioning and visual question answering~\cite{hu2025rsgpt,li2024hrvqa,wang2024earthvqa,zi2025rsvlm} to broader evaluations covering visual grounding, geospatial understanding, and complex reasoning~\cite{li2024vrsbench,danish2025geobench,an2025choice,wang2025xlrsbench}.
Geo3DVQA, for example, evaluates height-aware 3D geospatial reasoning from RGB-only aerial imagery~\cite{tsujimoto2026geo3dvqa}, so the geometry must be inferred from 2D appearance. SpatialSky-Bench evaluates spatial intelligence in UAV scenarios through distance, height, and navigation judgments, but does not provide registered height fields as evidence~\cite{zhang2026spatialsky}. GeoHeight-Bench centers on height-aware reasoning directly, providing relative-height and terrain-aware evaluations together with GeoHeightChat, a height-aware remote-sensing LMM baseline that augments optical models with implicitly injected height features~\cite{hu2026geoheight}. Broad remote-sensing benchmarks~\cite{danish2025geobench,an2025choice,wang2025xlrsbench} evaluate general remote-sensing understanding without explicit physical evidence.
In contrast, VertiCue-Bench provides pixel-aligned continuous height fields as explicit physical evidence and isolates whether models can perceive, ground, and utilize them, rather than asking models to infer geometry from appearance alone.

\subsection{Diagnostic and Intervention-based Evaluation}

Conventional benchmark accuracy cannot distinguish whether a correct prediction results from genuine physical-evidence use or from appearance priors and interface shortcuts.
Diagnostic and intervention-based evaluation provides a complementary perspective. Prior work has used carefully constructed visual probes to expose perception failures of MLLMs~\cite{tong2024mmvp}, polling-style tests to isolate object hallucination~\cite{li2023pope}, and counterfactual or compositional designs to test whether answers genuinely depend on the underlying evidence rather than on biases~\cite{niu2021counterfactualvqa}.
Following this spirit, our benchmark progressively reduces upstream uncertainty through Raw Visual, Tool-assisted, and Oracle Text conditions, and tests whether decisions change when the underlying physical condition changes.

In contrast to prior remote-sensing MLLM benchmarks, which primarily evaluate semantic understanding, grounding, or geospatial reasoning from optical imagery, our goal is to determine whether an MLLM can transform an explicit physical measurement into a decision-relevant constraint. This requires distinguishing three failure modes, namely physical evidence acquisition, spatial co-registration, and downstream utilization. VertiCue-Bench therefore combines pixel-aligned continuous height fields with a stage-wise task taxonomy and controlled representation interventions, and further uses counterfactual perturbations to test whether model decisions genuinely depend on vertical structure.

\section{Cognitive Framework for Vertical Structure Utilization}
\label{sec:framework}

The central challenge of vertical-structure reasoning is not merely whether height information is available, but whether physical evidence can be progressively transformed from a raw representation into information that can effectively support semantic decisions. We therefore formulate vertical-structure utilization as a three-stage evidence-utilization framework consisting of Perception, Grounding, and Utilization.

Let an evaluation item contain an RGB image $R$, an aligned physical height field $H$, a question $Q$, and answer options $O$. The required evidence flow is abstracted as
\begin{align}
    z_P &= \mathcal{P}(H,Q),\\
    z_G &= \mathcal{G}(R,z_P,Q),\\
    \hat{y} &= \mathcal{U}(R,z_G,Q,O),
\end{align}
where $\mathcal{P}$ transforms the available height representation into task-relevant physical measurements $z_P$, $\mathcal{G}$ binds these measurements to their corresponding visual entities to obtain grounded physical evidence $z_G$, and $\mathcal{U}$ converts the grounded evidence into constraints for the final decision $\hat{y}$. In this formulation, vertical evidence changes its functional role across the three stages, moving from a physical representation to an explicit measurement and then to an entity-grounded constraint for decision-making.

\subsection{Stage I: Perception}

Perception defines the model's ability to acquire task-relevant physical quantities from the available height representation. The source may be pixel-rendered CHM data or structured values returned through an external interface, while the resulting physical quantities may include absolute height, relative height, height difference, regional statistics, or structural variation. Formally, $\mathcal{P}$ represents the transformation from a raw height representation $H$ to explicit physical evidence $z_P$. At this stage, the requirement is to correctly recover what the physical measurement is, without requiring knowledge of which visual entity it belongs to.

\subsection{Stage II: Grounding}

Grounding defines the model's ability to spatially co-register the acquired physical measurements with their corresponding 2D visual entities. Given $z_P$, $\mathcal{G}$ establishes spatial associations between geometric quantities and target objects or regions in $R$, either in the pixel coordinate system or through an implicit spatial representation. This association can be bidirectional, from a visual entity to its physical measurement or from a geometric pattern to its corresponding visual entity. Grounding therefore transforms an isolated physical measurement into an entity-grounded physical fact, specifying not only what the measurement is, but also where and to which entity it belongs.

\subsection{Stage III: Utilization}

Utilization defines the model's ability to convert spatially grounded physical evidence into effective decision constraints and jointly reason over these constraints with semantic priors derived from 2D appearance. Given $z_G$, $\mathcal{U}$ determines whether the geometric evidence is properly incorporated into the final semantic or spatial decision. Under spectral ambiguity, successful utilization requires jointly reasoning over grounded vertical evidence and 2D appearance semantics to reach the correct semantic decision that resolves the ambiguity.

\subsection{Representation Intervention Spectrum}

The three-stage framework further motivates a Representation Intervention Spectrum that exposes the same underlying height evidence through Raw Visual, Tool-assisted, and Oracle Text interfaces. These interfaces progressively remove upstream barriers in the Perception--Grounding--Utilization chain, enabling stage-wise diagnosis.

In the Raw Visual passive pathway, the CHM is presented through pixel rendering and color mapping together with the corresponding RGB imagery, so the model must independently recover physical height measurements, maintain their spatial correspondence with visual entities, and use the resulting evidence for decision-making. Raw Visual therefore preserves all three requirements of Perception, Grounding, and Utilization.

In the Tool-assisted active interactive pathway, the model can actively query predefined point-, box-, or circle-based CHM functions to obtain structured height values or regional statistics. By directly returning physical measurements, Tool-assisted access substantially reduces the burden of decoding height from the visual rendering and thus largely externalizes the Perception stage. The model must nevertheless determine the appropriate spatial location to query, associate the returned measurements with the target entities, and use them in the final decision. Grounding and Utilization therefore remain required.

In the Oracle Text privileged description pathway, the Oracle interface directly provides task-relevant, entity-associated height facts in textual form without revealing the correct answer. It therefore substantially reduces the burden of both visual height decoding and explicit spatial association, effectively externalizing Perception and Grounding while retaining the requirement to convert the supplied physical facts into decision constraints. Oracle Text consequently provides a privileged condition for primarily probing the remaining Utilization requirement.

Accordingly, the three interfaces play distinct diagnostic roles. Raw Visual preserves all three stages ($\mathcal{P}+\mathcal{G}+\mathcal{U}$), Tool-assisted primarily externalizes $\mathcal{P}$, and Oracle Text further externalizes $\mathcal{P}+\mathcal{G}$. Their step-wise comparison provides the operational basis for diagnosing where vertical physical evidence fails to propagate toward semantic decision-making.

\subsection{Task--Intervention Mapping}
\label{sec:mapping}

To make the connection among the framework, the benchmark, and the experiments explicit, Table~\ref{tab:pgu_mapping} summarizes, for each stage, its core question, the tasks that predominantly exercise it, and the diagnostic intervention used to probe it.

\begin{table}[t]
  \centering
  \scriptsize
  \setlength{\tabcolsep}{3.0pt}
\caption{
  P--G--U, task, and intervention mapping.
  ``Dominant stage'' indicates the stage most directly exercised by the corresponding tasks rather than an exclusive instantiation.}
  \label{tab:pgu_mapping}
  {\renewcommand{\tabularxcolumn}[1]{m{#1}}%
  \begin{tabularx}{\columnwidth}{@{}>{\raggedright\arraybackslash}m{0.13\columnwidth} >{\raggedright\arraybackslash}X >{\raggedright\arraybackslash}X >{\raggedright\arraybackslash}X >{\raggedright\arraybackslash}X@{}}
    \toprule
    \textbf{Stage} & \textbf{Core question} & \textbf{Main tasks} & \textbf{Privileged condition} & \textbf{Diagnostic evidence}\\
    \midrule
    Perception & Can the model recover physical quantities? & L1--L3 & Tool & Raw$\rightarrow$Tool recovery \\
    Grounding & Can it bind measurements to entities? & L4; spatial components of L2--L3 & Oracle & Tool$\rightarrow$Oracle recovery \\
    Utilization & Can height alter the decision? & L5 & Oracle & Oracle residual; counterfactual response; semantic--geometric conflict \\
    \bottomrule
  \end{tabularx}
  }
\end{table}

We use dominant stage rather than assuming that each task exclusively instantiates a single stage.
L2--L3, for example, may require both physical measurement and spatial organization, while L5 inherits the requirements of upstream stages.

\section{Design and Implementation of VertiCue-Bench}
\label{sec:benchmark}

\begin{figure}[t]
  \centering
  \includegraphics[width=\textwidth]{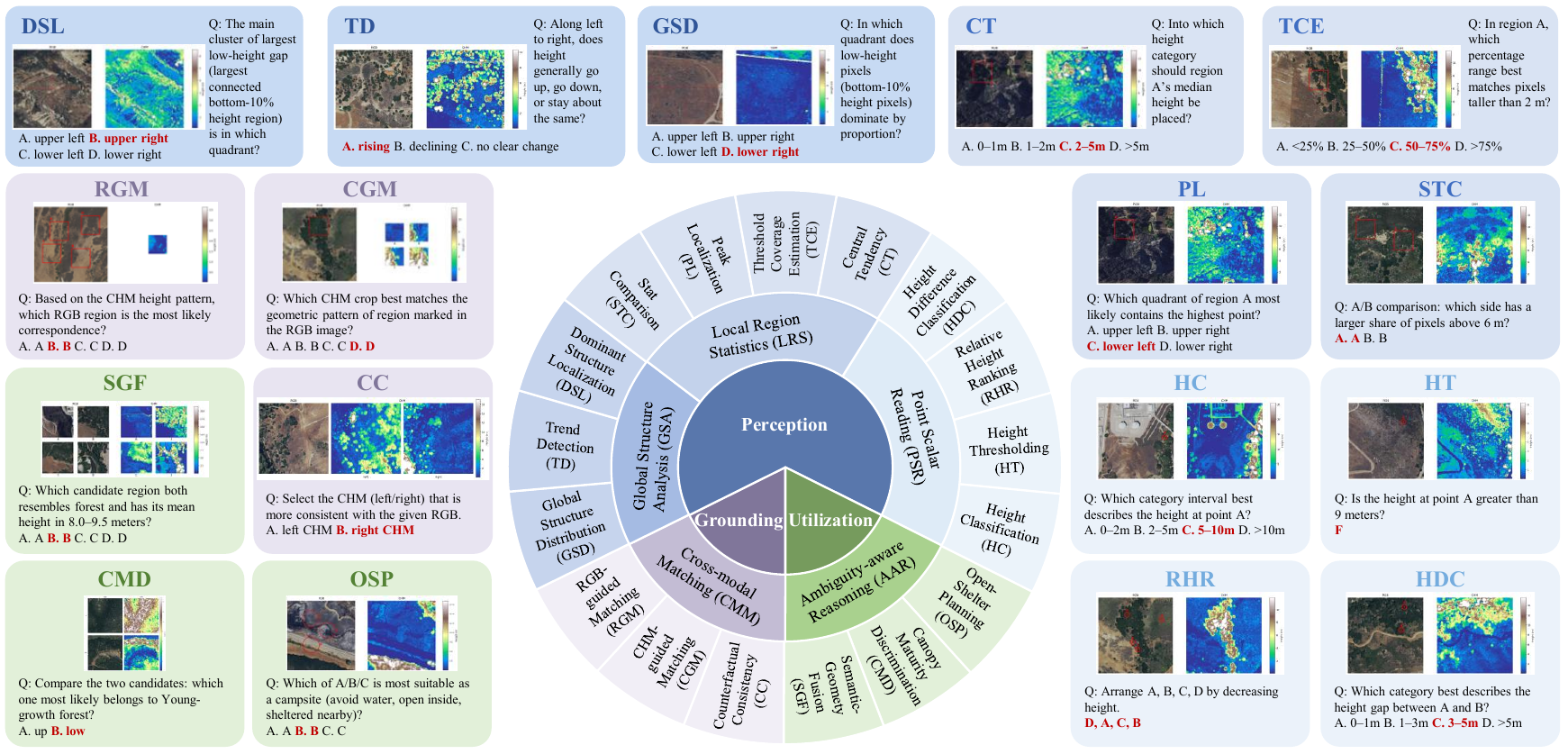}
  \caption{Five-Level task taxonomy of VertiCue-Bench.}
  \label{fig:hierarchy}
\end{figure}

Fig.~\ref{fig:hierarchy} overviews the five-level task hierarchy that VertiCue-Bench is built around; the following subsections describe how the physical evidence is selected, why the tasks are designed in this way, how the data are constructed, and what the resulting benchmark contains.

\subsection{Selection of Physical Evidence: Why CHM?}

To serve as reliable physical evidence for vertical-structure reasoning, the selected representation should provide direct physical measurements, metric interpretability, and continuous spatial support. We therefore adopt the CHM as the source of vertical evidence. CHMs quantify canopy height above the underlying terrain in metric units and preserve its spatial distribution, providing a physically grounded representation of vegetation vertical structure. As a standard LiDAR-derived product, CHM has been widely used in remote sensing for forest inventory, structural characterization, and vegetation analysis~\cite{Balestra2024,Allred2025,yao2025mapping}.

CHM is particularly suited to the spectral ambiguity considered in this work. Spectrally similar regions, such as shrubs, regenerating woodland, and mature forest, can exhibit substantially different canopy heights, while properties such as openness and surrounding shelter are also directly reflected by vertical structure. CHM therefore provides an established and physically interpretable source of evidence for distinguishing natural scenes that cannot be reliably resolved from 2D optical appearance alone.

In this role, CHM-derived height serves throughout VertiCue-Bench as metric geometric evidence or a geometric constraint, rather than as semantic ground truth itself.

\subsection{Task Design Principles}

Five principles guide the design of every task in VertiCue-Bench.

\begin{itemize}
    \setlength{\itemsep}{1pt}
    \setlength{\parskip}{0pt}
    \setlength{\parsep}{0pt}
    \setlength{\topsep}{2pt}

    \item Physical measurability: each task must involve explicit physical quantities derived from the CHM as part of its evidence or decision criteria.

    \item Spatial specificity: height evidence must be associated with a well-defined point, region, candidate, or spatial structure relevant to the task.

    \item Semantic ambiguity: especially at L5, RGB appearance alone should not reliably determine the answer, so that vertical evidence is genuinely decision-critical.

    \item Decision sensitivity: in utilization tasks, changing the relevant height quantity must be able to change the ground truth.

    \item Shortcut resistance: cases solvable from colorbar ranges, obvious extrema, answer imbalance, unstable thresholds, or trivial semantic cues are filtered out.
\end{itemize}

These principles determine the progression from direct physical measurement to spatially grounded and ambiguity-aware decision-making, resulting in the five-level taxonomy below.

\begin{figure}[t]
  \centering
  \includegraphics[width=\textwidth]{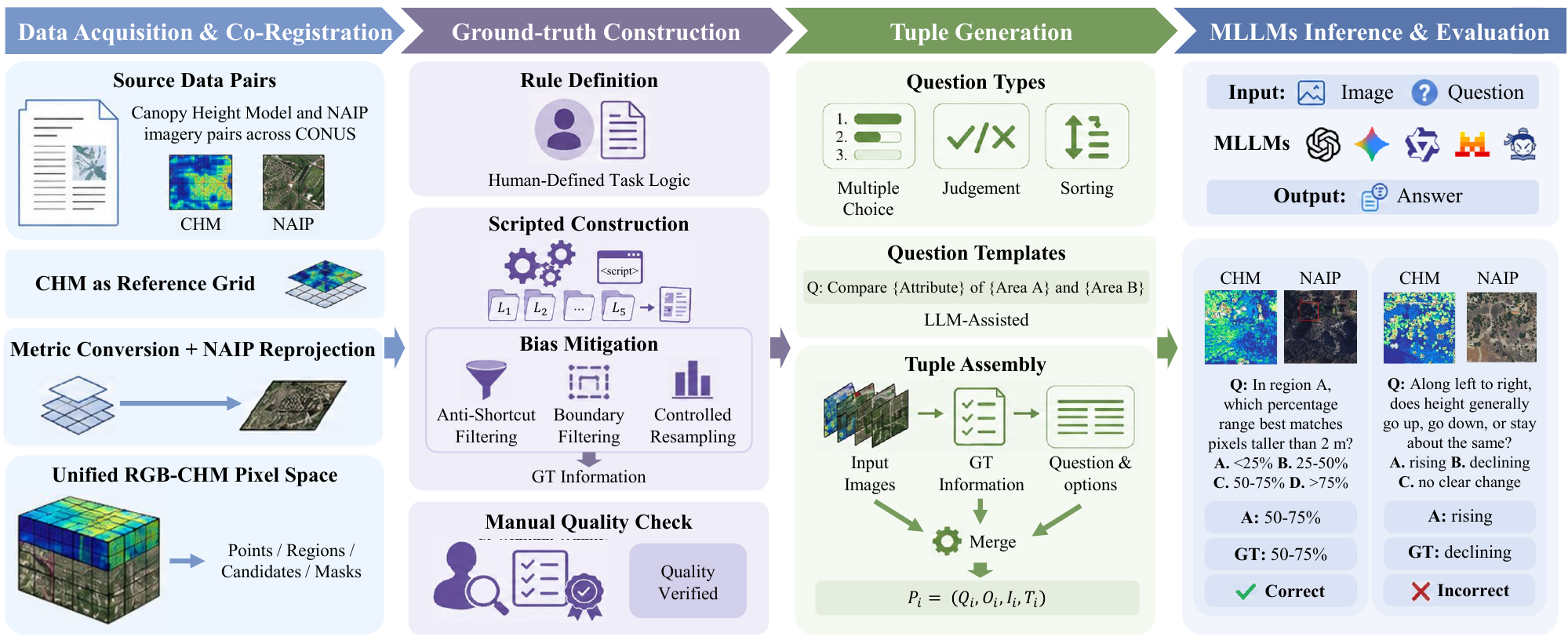}
  \caption{Construction and evaluation pipeline of VertiCue-Bench.}
  \label{fig:pipeline_main}
\end{figure}

\subsection{Five-Level Task Taxonomy}
\label{sec:taxonomy}

As shown in Fig.~\ref{fig:hierarchy}, VertiCue-Bench comprises 17 fine-grained tasks, organized into five levels and mapped to the Perception--Grounding--Utilization framework according to their dominant evidence requirements.
Table~\ref{tab:taxonomy} lists the complete taxonomy; the paragraphs below explain, for each level, the capability it targets, its input and operation, and why the level matters.

\begin{table}[t]
  \centering
  \scriptsize
  \setlength{\tabcolsep}{2.5pt}
\caption{
  VertiCue-Bench task taxonomy.
  Each task is designed to test one or more capabilities in the P--G--U chain; the mapping to dominant stages is summarized in Table~\ref{tab:pgu_mapping}.}
  \label{tab:taxonomy}
  \begin{tabularx}{\columnwidth}{@{}>{\raggedright\arraybackslash}p{0.08\columnwidth} >{\raggedright\arraybackslash}p{0.18\columnwidth} >{\raggedright\arraybackslash}X@{}}
    \toprule
    \textbf{Level} & \textbf{Task} & \textbf{What it tests}\\
    \midrule
    L1 & HC & Point-wise height classification\\
    L1 & HT & Height threshold reasoning\\
    L1 & RHR & Relative height ranking\\
    L1 & HDC & Height difference classification\\
    L2 & CT & Region-level statistics\\
    L2 & TCE & Threshold coverage estimation\\
    L2 & PL & Peak localization\\
    L2 & STC & Statistical comparison across regions\\
    L3 & DSL & Dominant structure localization\\
    L3 & TD & Trend detection\\
    L3 & GSD & Global structure distribution comparison\\
    L4 & RGM & CHM-pattern matching to RGB candidates\\
    L4 & CGM & RGB-region matching to CHM candidates\\
    L4 & CC & RGB--CHM consistency under counterfactual swaps\\
    L5 & SGF & Semantic--height constraint integration\\
    L5 & CMD & Appearance--height conflict in maturity discrimination\\
    L5 & OSP & Multi-constraint spatial planning\\
    \bottomrule
  \end{tabularx}
\end{table}

L1 covers point-level scalar reading and belongs to the Perception stage.
Given a CHM, one or more query points marked in the co-registered RGB image, and a quantitative query, L1 evaluates whether the model can recover and compare point-level physical height information.
Height Classification (HC), Height Thresholding (HT), Relative Height Ranking (RHR), and Height Difference Classification (HDC) respectively require height categorization, threshold judgment, relative ordering, and difference classification.
Because these tasks do not require regional spatial organization or RGB semantic grounding, L1 provides the most basic test of the Perception stage.

L2 and L3 broaden the evaluation to local region statistics and global structure analysis, spanning Stages I and II (Perception--Grounding).
L2 extends the evaluation from point-wise scalars to predefined CHM regions, requiring models to aggregate and organize regional height evidence through Central Tendency (CT), Threshold Coverage Estimation (TCE), Peak Localization (PL), and Stat Comparison (STC).
L3 further expands the spatial scope to the full-scene CHM, where Dominant Structure Localization (DSL), Trend Detection (TD), and Global Structure Distribution (GSD) probe the ability to identify dominant structures, global trends, and spatial distributions.
Together, L2--L3 move beyond direct value reading toward the spatial organization of physical evidence, naturally bridging Perception and subsequent Grounding.

L4 covers cross-modal matching and belongs to the Grounding stage.
L4 introduces RGB regions and CHM patterns as cross-modal inputs and evaluates whether the model can establish reliable correspondence between them rather than merely interpret the height field itself.
RGB-guided Matching (RGM) requires matching a CHM pattern to the most compatible RGB candidate, CHM-guided Matching (CGM) performs the reverse association, and Counterfactual Consistency (CC) asks the model to discriminate between real and counterfactually swapped CHM evidence for the same RGB input.
These tasks directly target Grounding by testing whether height-derived geometric structures can be associated with the appropriate visual entities.

L5 covers ambiguity-aware reasoning and belongs to the Utilization stage.
L5 focuses on appearance-ambiguous RGB candidates paired with aligned CHMs and task-specific decision conditions, testing whether models can jointly integrate the aligned height evidence with semantic or spatial requirements for the final decision.
Semantic--Geometry Fusion (SGF) requires the same candidate to satisfy both an appearance-based semantic condition and an explicit height constraint; Canopy Maturity Discrimination (CMD) uses height differences to distinguish visually similar mature high-canopy forest from lower shrubs or young stands; and Open--Shelter Planning (OSP) combines openness, surrounding shelter, and other spatial conditions in a multi-constraint decision.
At this level, height is no longer merely a quantity to be predicted or matched, but becomes a decision constraint whose satisfaction can alter the final answer, directly probing the Utilization stage.

\begin{figure}[t]
  \centering
  \subfloat[Task and answer counts]{%
    \label{fig:samples_answers}%
    \includegraphics[width=.35\textwidth]{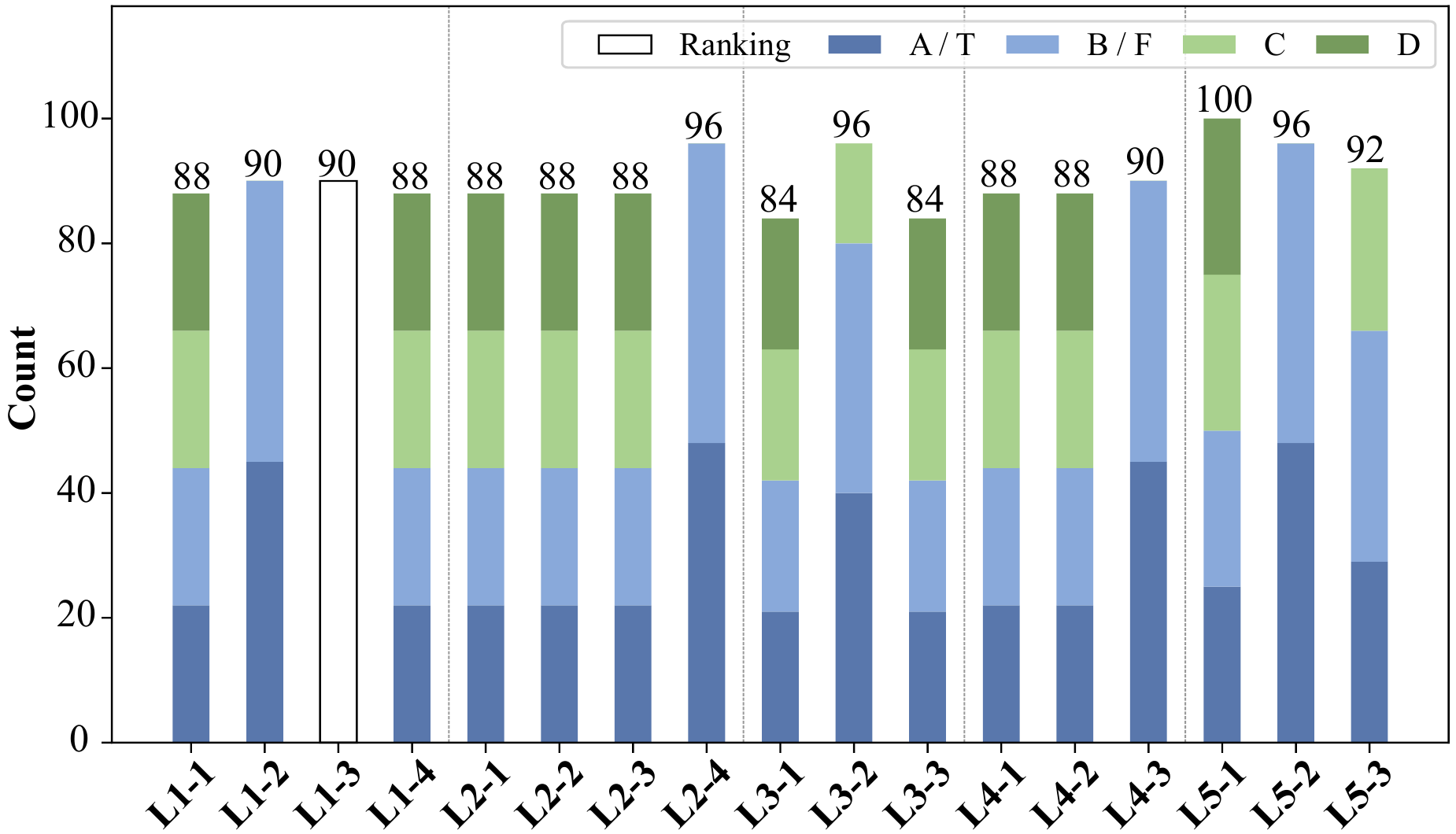}
  }\hfill
  \subfloat[Land-cover composition]{%
    \label{fig:semantic_category_pie}%
    \includegraphics[width=.27\textwidth]{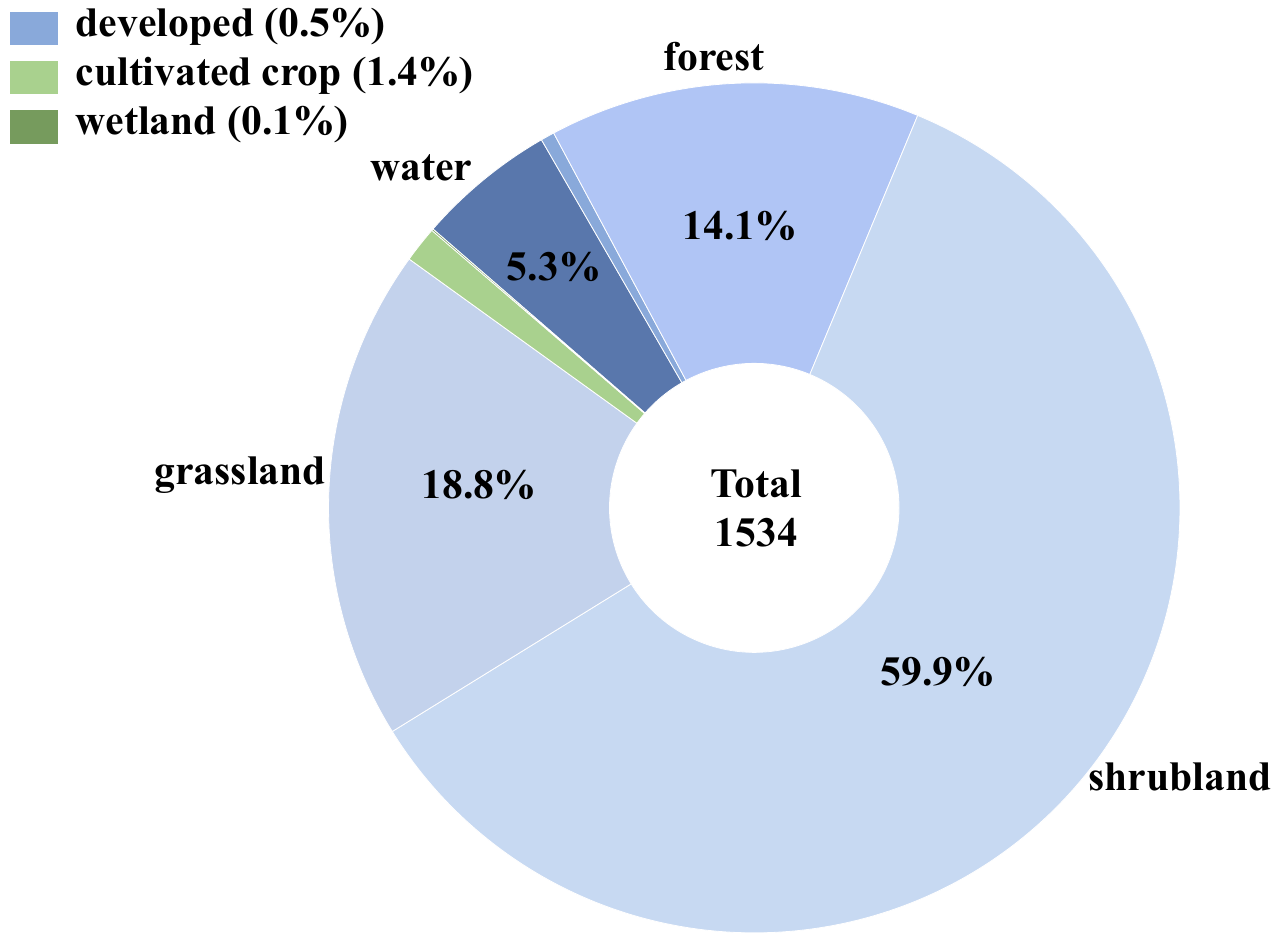}
  }\hfill
  \subfloat[Question vocabulary]{%
    \label{fig:question_wordcloud}%
    \includegraphics[width=.33\textwidth]{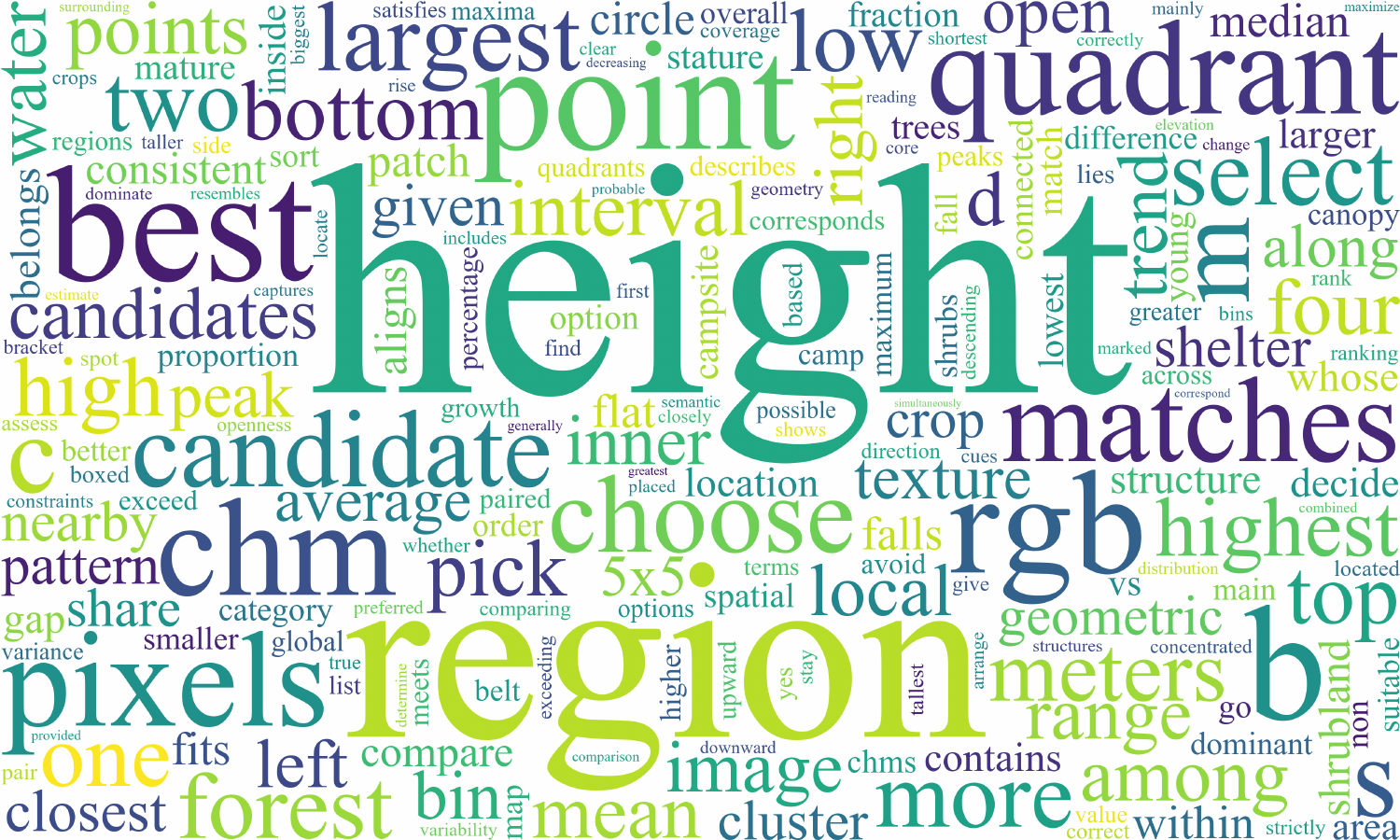}
  }
  \caption{Overview of the dataset statistics.
  (a) Sample counts and answer distribution for each task, with the x-axis codes indexing the tasks listed in Table~\ref{tab:taxonomy}.
  (b) Semantic category distribution.
  (c) Word cloud of the questions.}
  \label{fig:dataset_overview}
\end{figure}

\subsection{Data Acquisition and Pixel-Level Co-Registration}

As shown in Fig.~\ref{fig:pipeline_main}, VertiCue-Bench is built upon the public Canopy height model and NAIP imagery pairs across CONUS dataset~\cite{Allred2025}. The source data pair 1-m CHM chips derived from USGS 3DEP airborne LiDAR with temporally aligned USDA NAIP aerial imagery, covering diverse ecoregions and land-cover classes across the conterminous United States. Approximately 11,000 CHM--NAIP pairs are used as the initial candidate pool.

Although the source pairs are geographically matched, their native spatial resolutions and pixel grids are not identical. We therefore use the CHM grid as the reference coordinate system for pixel-level co-registration. Raw CHM values are converted to metric heights, and NAIP imagery is reprojected onto the CHM grid. Continuous image bands are resampled using bilinear interpolation, whereas the validity mask is resampled using nearest-neighbor interpolation to preserve valid-data boundaries. Only pixels valid in both modalities are retained. All subsequent points, regions, candidate crops, and masks are generated in this unified pixel space, ensuring that corresponding RGB and CHM pixels refer to the same ground footprint.

\subsection{Ground-Truth Construction and Bias Mitigation}

As shown in Fig.~\ref{fig:pipeline_main}, VertiCue-Bench constructs and instantiates its annotations through rule definition, scripted construction, bias mitigation, manual verification, and tuple generation.

For each task, we first define the physical quantities, thresholds, spatial relations, semantic--geometric conditions, and answer rules required to determine the correct answer.
These components are manually specified before language generation, ensuring that ground-truth labels are fully determined by human-defined task logic.

For the L5 tasks, these predefined rules explicitly combine task-specific semantic and geometric conditions.
For SGF, a candidate is correct only when it simultaneously satisfies the target land-cover semantic condition and the corresponding class-specific height range, which is determined from class-specific CHM height statistics; for example, when the target class is forest, the candidate must also have a mean canopy height within 8.0--9.5~m.
For CMD, the land-cover semantic class and the height-derived high/low condition are combined through a predefined decision rule; for example, a forest candidate satisfying the high-height condition is labeled mature high-canopy forest, whereas forest and shrubland candidates satisfying the low-height condition are labeled young forest and low-stature shrubland, respectively.
For OSP, water acts as a hard exclusion, while inner-region openness and surrounding shelter are quantified from spatial statistics; for example, a preferred site should avoid water, remain relatively open internally, and have sufficiently high surrounding structure for shelter.

The predefined rules are then applied to the registered RGB--CHM data to generate the GT Information. L1--L3 derive answers from CHM measurements and spatial statistics; L4 uses RGB--CHM spatial correspondence; and L5 combines semantic conditions with vertical-structure constraints. Three controls are applied to mitigate potential bias. Anti-shortcut filtering removes cases solvable from unintended visual cues such as colorbar ranges or obvious extrema; boundary filtering imposes safety margins around numerical thresholds, candidate-matching margins, and semantic--geometric decision boundaries to exclude unstable cases; and controlled resampling limits overrepresented answer categories and resamples valid candidates to maintain approximately balanced answer distributions within each task.

The generated GT records are subsequently verified for RGB--CHM registration, question--evidence consistency, answer uniqueness, and visualization quality. A stratified manual audit of 300 randomly sampled instances finds 295 (98.3\%) fully consistent with the rendered evidence and stored GT on the first pass; the remaining cases are corrected or removed.

VertiCue-Bench adopts three question types (Multiple Choice, Judgement, and Sorting) to facilitate unified quantitative evaluation across tasks. Once the ground truth is fixed, the task-specific question templates are instantiated for each GT record. Large language model (LLM) assistance is used for linguistic realization and question-template diversification. In Tuple Assembly, the question, answer options, input images, and corresponding GT information are combined into the final tuple,
\begin{equation}
    P_i=(Q_i,O_i,I_i,T_i),
\end{equation}
where $Q_i$ denotes the question, $O_i$ the answer space, $I_i$ the input representation, and $T_i$ the ground-truth answer. The resulting tuples constitute standardized evaluation instances for subsequent MLLM inference and benchmark evaluation.

\subsection{Dataset Statistics}

VertiCue-Bench contains 1,534 standardized evaluation instances spanning 17 task types. VertiCue-Bench is designed as a diagnostic evaluation suite rather than a training corpus; accordingly, it prioritizes coverage of distinct capability conditions and controlled interventions over sheer scale. As shown in Fig.~\ref{fig:dataset_overview}\subref{fig:samples_answers}, the sub-tasks are generally balanced in both sample size and answer distribution, which alleviates potential bias from label imbalance. As shown in Fig.~\ref{fig:dataset_overview}\subref{fig:semantic_category_pie} 
and Fig.~\ref{fig:dataset_overview}\subref{fig:question_wordcloud}, 
the benchmark instances cover diverse land-cover semantics, while the question vocabulary simultaneously emphasizes geometric and semantic concepts such as height and forest. Together, these observations indicate that VertiCue-Bench is deliberately designed to probe the geometry--semantics gap in remote sensing, rather than focusing on pure geometric reading or isolated semantic recognition.

\section{Benchmarking MLLMs on Passive Visual Channel}
\label{sec:passive}
\subsection{Experimental Setup}
We evaluate the full set of 1,534 instances under the canonical passive visual channel, where each model receives a text-first prompt with a side-by-side RGB+CHM composite and a metric colorbar; this canonical presentation is also the baseline condition in the presentation-robustness analysis in Section~\ref{sec:diagnostics}. An RGB-only condition is additionally evaluated on L1--L3 and L5 as a modality control, while L4 is excluded because its cross-modal matching tasks intrinsically require both RGB and CHM. We evaluate 10 MLLMs from three model families. Proprietary models include GPT-5-mini~\cite{singh2025openaigpt5card} and Gemini-3-Flash; general-purpose open-source models include Qwen3.5-9B and Qwen3.5-27B~\cite{qwen35blog}, InternVL3.5-8B and InternVL3.5-14B~\cite{wang2025internvl35advancingopensourcemultimodal}, Mistral-Small-3.2-24B~\cite{MistralAI}, and Gemma-3-12B~\cite{gemmateam2025gemma3technicalreport}; and remote-sensing-specialized models include RemoteReasoner~\cite{yao2026remotereasoner} and RSThinker~\cite{liu2026rsthinker}. All models are evaluated through their native inference interfaces using a unified zero-shot strategy with deterministic decoding ($temperature=0$), and Accuracy is used as the evaluation metric.
Tool interface availability is model-dependent; models without a compatible tool interface are evaluated only under Raw and Oracle conditions (Section~\ref{sec:diagnostics}).

\subsection{Baseline Capability under Passive Visual Input}
\begin{table}[t]
  \centering
  \footnotesize
  \setlength{\tabcolsep}{4.6pt}
  \renewcommand{\arraystretch}{1.06}
  \caption{\textbf{Main results on VertiCue-Bench across different model families.} Accuracy is reported in percent under the canonical RGB+CHM passive visual channel.}
  \label{tab:main_results}
  \begin{tabular*}{\textwidth}{@{\extracolsep{\fill}}lccccc|cccc@{}}
    \toprule
    \multirow{2}{*}{\textbf{Models}} &
    \multicolumn{5}{c|}{\textbf{Perception and Grounding}} &
    \multicolumn{4}{c}{\textbf{Utilization}}\\
    \cmidrule(lr){2-6}\cmidrule(l){7-10}
    & \textbf{L1} & \textbf{L2} & \textbf{L3} & \textbf{L4} & \textbf{Overall}
    & \textbf{SGF} & \textbf{CMD} & \textbf{OSP} & \textbf{Overall}\\
    \midrule

    \rowcolor{orange!5}\multicolumn{10}{l}{\textit{\textbf{1. Proprietary Large Vision-Language Models}}}\\
    GPT-5-mini 
    & 46.6 & \uline{60.0} & \textbf{54.6} & 58.3 & \uline{54.7}
    & \textbf{42.0} & \uline{76.0} & \uline{85.9} & \textbf{67.4}\\
    
    Gemini-3-Flash 
    & 46.9 & \textbf{64.4} & 50.8 & \textbf{65.7} & \textbf{56.8}
    & 32.0 & \textbf{80.2} & \textbf{89.1} & \uline{66.3}\\

    \midrule
    \rowcolor{blue!5}\multicolumn{10}{l}{\textit{\textbf{2. Open-Source Large Vision-Language Models}}}\\
    Qwen3.5 (9B) 
    & \uline{48.6} & 53.3 & 48.9 & 55.3 & 51.4
    & \textbf{42.0} & 69.8 & 82.6 & 64.2\\

    Qwen3.5 (27B) 
    & \textbf{50.3} & 55.8 & \uline{51.1} & \uline{60.9} & 54.3
    & \uline{39.0} & 74.0 & 77.2 & 62.9\\

    InternVL3.5 (8B) 
    & 26.7 & 35.8 & 32.2 & 37.2 & 32.7
    & 37.0 & 55.2 & 62.0 & 51.0\\

    InternVL3.5 (14B) 
    & 34.6 & 44.7 & 34.1 & 36.8 & 37.9
    & \textbf{42.0} & 62.5 & 58.7 & 54.2\\

    Mistral-Small-3.2 (24B) 
    & 38.2 & 35.8 & 34.1 & 38.7 & 36.8
    & 33.0 & 51.0 & 46.7 & 43.4\\

    Gemma-3 (12B) 
    & 32.9 & 36.7 & 28.4 & 38.4 & 34.2
    & 28.0 & 51.0 & 56.5 & 44.8\\

    \midrule
    \rowcolor{purple!5}\multicolumn{10}{l}{\textit{\textbf{3. Remote Sensing Specialized Models}}}\\

    RemoteReasoner 
    & 30.1 & 31.9 & 25.4 & 30.5 & 30.7
    & 35.0 & 60.4 & 27.2 & 41.0\\

    RSThinker 
    & 24.9 & 32.2 & 23.5 & 33.5 & 28.6
    & 18.0 & 46.9 & 23.9 & 29.5\\

    \bottomrule
  \end{tabular*}
\end{table}

Table~\ref{tab:main_results} reveals a clear performance hierarchy across model families under the canonical RGB+CHM passive visual channel. Proprietary models form the strongest tier on both Perception--Grounding and Utilization, general-purpose open-source models occupy the intermediate tier, and remote-sensing-specialized models lag substantially behind. The best Perception--Grounding scores across the three families are 56.8\%, 54.3\%, and 30.7\%, respectively, while their best Utilization scores are 67.4\%, 64.2\%, and 41.0\%. This consistent stratification indicates that remote-sensing specialization alone does not naturally translate into stronger processing of explicit height evidence.

Meanwhile, the stage-wise performance hierarchy is not invariant across capability stages. Gemini-3-Flash achieves the highest Perception--Grounding Overall score (56.8\%), whereas GPT-5-mini leads on Utilization Overall (67.4\%). Similarly, scaling Qwen3.5 from 9B to 27B improves Perception--Grounding Overall from 51.4\% to 54.3\%, but reduces Utilization Overall from 64.2\% to 62.9\%. This stage-dependent divergence suggests that stronger acquisition and spatial processing of height evidence do not necessarily entail stronger downstream utilization.

\begin{figure}[t]
  \centering
  \includegraphics[width=\columnwidth]{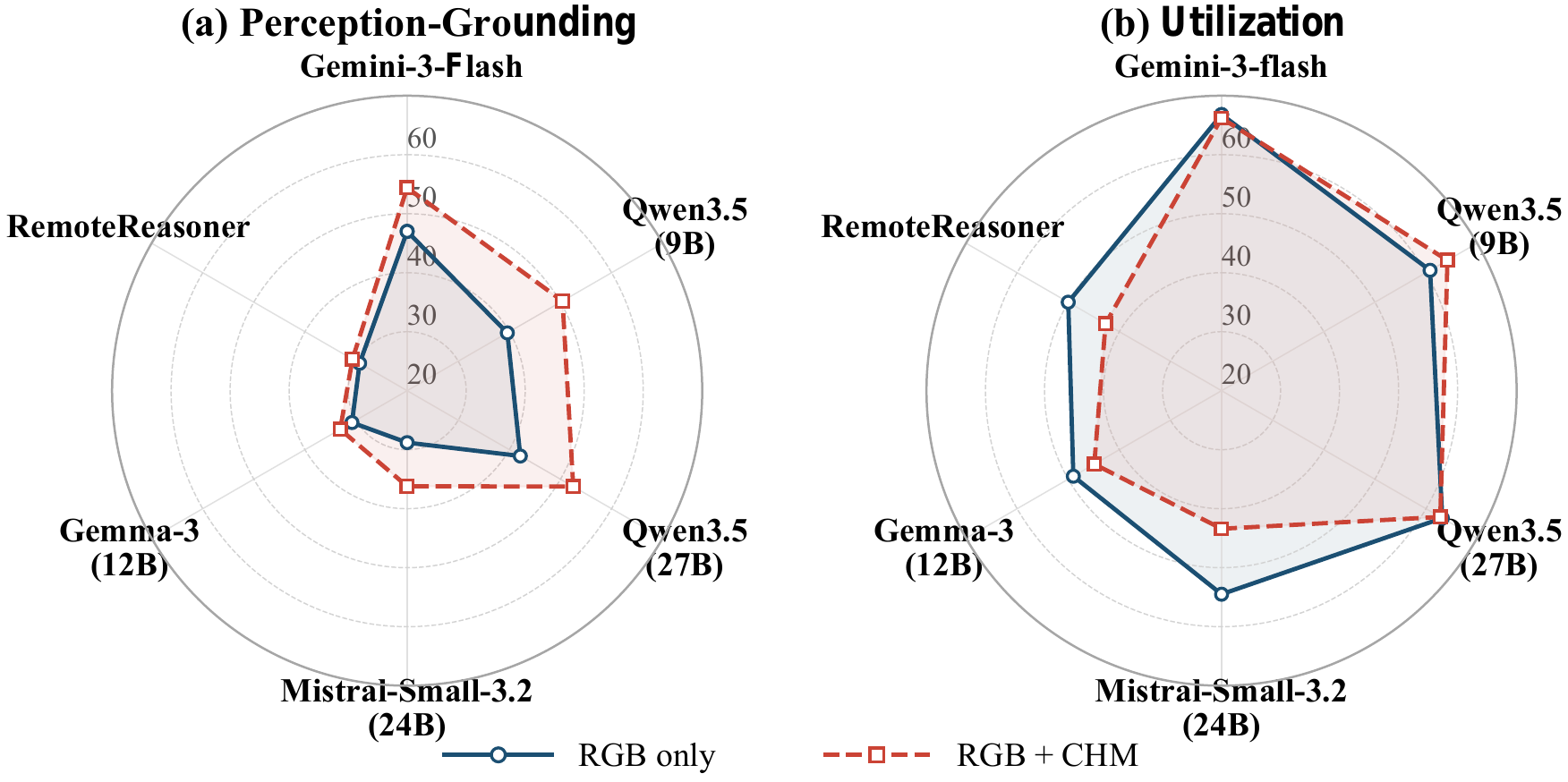}
  \caption{RGB-only versus RGB+CHM performance for six representative models. (a) Perception--Grounding aggregates L1--L3; (b) Utilization corresponds to L5.}
  \label{fig:modality_ablation}
\end{figure}

Fig.~\ref{fig:modality_ablation} further reveals a clear geometry-to-semantics gap in current MLLMs. Adding CHM consistently improves Perception--Grounding by enabling models to extract and organize height-related cues, yet it does not reliably enhance Utilization of these cues. Across models, Utilization performance may improve only marginally, remain nearly unchanged, or even degrade relative to RGB-only inputs. Even when gains are observed, they are typically much smaller than the corresponding improvements in Perception--Grounding. Thus, the additional geometric information provided by CHM does not consistently propagate into downstream semantic decisions.

\subsection{Can Models Acquire Vertical Evidence?}

The passive-channel results demonstrate that current MLLMs can already extract useful vertical-structure cues under Raw CHM, indicating that they are not fundamentally incapable of perceiving height evidence.
However, the overall Perception--Grounding performance remains limited, showing that physical evidence acquisition and spatial organization are not yet sufficiently reliable.
The primary limitation at this stage is therefore not complete blindness to height, but the reliability of physical-evidence access and spatial organization.

\begin{figure}[t]
  \centering
  \includegraphics[width=1\textwidth]{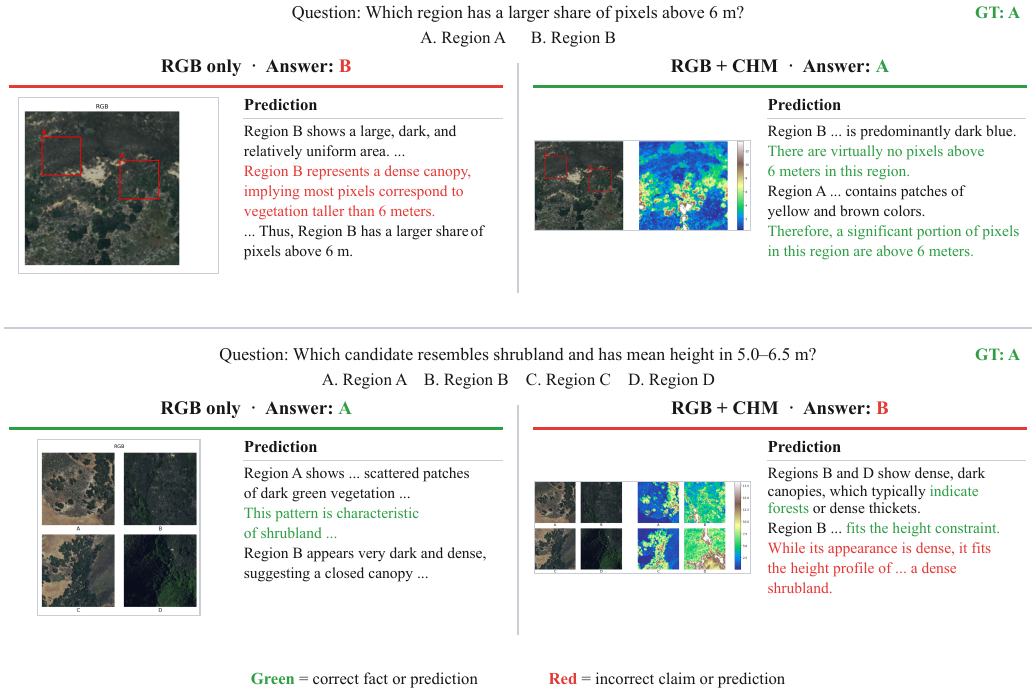}
  \caption{Representative cases of the vertical structure utilization gap. The top panel shows CHM correcting an L2-4 height-perception error. The bottom panel shows an L5-1 SGF case where incorrect semantic--geometry fusion turns a correct RGB-only prediction into an incorrect RGB+CHM prediction.}
  \label{fig:qualitative_modality}
\end{figure}

\subsection{Can Models Ground Vertical Evidence?}

The second question concerns spatial binding, i.e., whether the model can associate height-derived physical evidence with the correct visual entity.
Direct evidence comes from the L4 cross-modal matching tasks in Table~\ref{tab:main_results}, where even the strongest model reaches only 65.7\% on L4 Overall and remote-sensing-specialized models fall below 34\%, indicating that RGB--CHM correspondence remains limited.

\subsection{Can Models Use Vertical Evidence in Semantic Decisions?}

Even when models exhibit some ability to access and ground height evidence, the more consequential limitation arises in cross-stage evidence propagation.
The stage-dependent performance divergence and asymmetric CHM gains observed in Section~\ref{sec:passive} indicate that accessing height evidence does not necessarily imply that it can be transformed into effective evidence governing the final semantic decision.
We refer to this failure of low-level geometric capability to reliably propagate toward higher-level decision-making as the Vertical Structure Utilization Gap, which Section~\ref{sec:diagnostics} formalizes and diagnoses in detail.
Fig.~\ref{fig:qualitative_modality} provides a sample-level visualization of this vertical structure utilization gap.

Among all downstream reasoning tasks reported in Table~\ref{tab:main_results}, SGF exhibits the most pronounced difficulty, ranking as the lowest-scoring downstream reasoning task for 9 of the 10 evaluated models.
For example, Gemini-3-Flash achieves only 32.0\% on SGF, compared with 80.2\% on CMD and 89.1\% on OSP.
Unlike the other tasks, SGF requires the same candidate to simultaneously satisfy an appearance-based semantic condition and an explicit height-range constraint.
This result therefore further localizes the Utilization bottleneck to jointly integrating vertical geometric evidence with semantic information in the final decision.
SGF therefore provides a direct operational test of whether accessible physical evidence can effectively constrain the final decision.

\section{Diagnosing the Vertical Structure Utilization Gap}
\label{sec:diagnostics}
\subsection{Step-wise Diagnostic Intervention: Raw vs. Tool vs. Oracle}

To progressively localize failures along the Perception--Grounding--Utilization chain, we evaluate the Representation Intervention Spectrum on the full 1,534-instance benchmark. Raw Visual preserves the complete evidence-utilization chain. 
Tool-assisted access directly returns structured CHM measurements through point-, box-, and circle-based queries, thereby removing much of the perceptual burden of decoding metric values from rendered colors and pixels, while still requiring models to determine where to query and how to associate the returned measurements. 
Oracle Text further provides task-relevant geometric height facts in textual form, substantially reducing the spatial localization and measurement-association burden.

The step-wise comparison is organized around three questions, in which Raw$\rightarrow$Tool asks whether physical-measurement acquisition is a bottleneck; Tool$\rightarrow$Oracle asks whether spatial grounding is a bottleneck; and the residual error under Oracle tests whether, even with explicit and entity-associated evidence, the model can use it as a decision constraint.

\begin{table}[t]
  \centering
  \scriptsize
  \setlength{\tabcolsep}{2.0pt}
  \renewcommand{\arraystretch}{1.04}

  \caption{Raw Visual, Tool-assisted, and Oracle Text evaluation on VertiCue-Bench.}
  \label{tab:rto_full}

  \begin{tabularx}{\columnwidth}{
    @{}
    >{\raggedright\arraybackslash}p{0.23\columnwidth}
    >{\centering\arraybackslash}p{0.105\columnwidth}
    *{5}{>{\centering\arraybackslash}X}
    >{\centering\arraybackslash}p{0.105\columnwidth}
    >{\centering\arraybackslash}p{0.075\columnwidth}
    @{}
  }

    \toprule

    \multirow{2}{*}{\textbf{Model}}
    & \multirow{2}{*}{\textbf{Setting}}
    & \multicolumn{5}{c}{\textbf{Accuracy (\%)}}
    & \multirow{2}{*}{\textbf{Overall}$\uparrow$}
    & \multirow{2}{*}{\textbf{$\Delta$}}
    \\

    \cmidrule(lr){3-7}

    &
    & \textbf{L1}
    & \textbf{L2}
    & \textbf{L3}
    & \textbf{L4}
    & \textbf{L5}
    &
    &
    \\

    \midrule

    & Raw
    & 46.6
    & 60.0
    & 54.6
    & 58.3
    & 67.4
    & 57.0
    & -- \\

    \rowcolor{toolbg}
    GPT-5-mini
    & Tool
    & 44.9
    & 63.6
    & 63.6
    & 50.4
    & \textbf{75.4}
    & \uline{59.2}
    & \textcolor{green!50!black}{+2.2} \\

    \rowcolor{oraclebg}
    & Oracle
    & 100.0
    & 99.2
    & 89.4
    & 55.3
    & \uline{68.8}
    & \textbf{84.4}
    & \textcolor{green!50!black}{+27.4} \\

    \midrule

    & Raw
    & 46.9
    & 64.4
    & 50.8
    & 65.7
    & 66.3
    & 58.6
    & -- \\

    \rowcolor{toolbg}
    Gemini-3-Flash
    & Tool
    & 58.4
    & 88.3
    & 66.7
    & 66.2
    & \textbf{80.2}
    & \uline{72.3}
    & \textcolor{green!50!black}{+13.7} \\

    \rowcolor{oraclebg}
    & Oracle
    & 99.7
    & 99.4
    & 90.5
    & 66.5
    & \uline{78.1}
    & \textbf{88.3}
    & \textcolor{green!50!black}{+29.7} \\

    \midrule

    & Raw
    & 48.6
    & 53.3
    & 48.9
    & 55.3
    & 64.2
    & 53.9
    & -- \\

    \rowcolor{toolbg}
    Qwen3.5 (9B)
    & Tool
    & 57.9
    & 72.8
    & 65.9
    & 71.4
    & \textbf{83.0}
    & \uline{69.8}
    & \textcolor{green!50!black}{+15.9} \\

    \rowcolor{oraclebg}
    & Oracle
    & 100.0
    & 99.2
    & 89.0
    & 68.1
    & \uline{78.8}
    & \textbf{88.4}
    & \textcolor{green!50!black}{+34.5} \\

    \midrule

    & Raw
    & 50.3
    & 55.8
    & 51.1
    & 60.9
    & 62.9
    & 55.9
    & -- \\

    \rowcolor{toolbg}
    Qwen3.5 (27B)
    & Tool
    & 48.5
    & 61.8
    & 57.4
    & 59.0
    & \textbf{77.1}
    & \uline{60.3}
    & \textcolor{green!50!black}{+4.4} \\

    \rowcolor{oraclebg}
    & Oracle
    & 100.0
    & 99.4
    & 87.5
    & 66.5
    & \uline{74.3}
    & \textbf{87.1}
    & \textcolor{green!50!black}{+31.2} \\

    \midrule

    & Raw
    & 26.7
    & 35.8
    & 32.2
    & 37.2
    & 51.0
    & 36.2
    & -- \\

    \rowcolor{toolbg}
    InternVL3.5 (8B)
    & Tool
    & 35.8
    & 43.9
    & 33.3
    & 44.4
    & \uline{51.0}
    & \uline{41.6}
    & \textcolor{green!50!black}{+5.4} \\

    \rowcolor{oraclebg}
    & Oracle
    & 89.0
    & 92.5
    & 58.0
    & 41.7
    & \textbf{62.9}
    & \textbf{71.4}
    & \textcolor{green!50!black}{+35.2} \\

    \midrule

    & Raw
    & 34.6
    & 44.7
    & 34.1
    & 36.8
    & 54.2
    & 40.9
    & -- \\

    \rowcolor{toolbg}
    InternVL3.5 (14B)
    & Tool
    & 36.2
    & 39.7
    & 29.6
    & 36.1
    & \uline{52.8}
    & \uline{39.0}
    & \textcolor{red!70!black}{-1.9} \\

    \rowcolor{oraclebg}
    & Oracle
    & 83.4
    & 97.5
    & 72.7
    & 38.3
    & \textbf{64.6}
    & \textbf{73.5}
    & \textcolor{green!50!black}{+32.6} \\

    \midrule

    & Raw
    & 38.2
    & 35.8
    & 34.1
    & 38.7
    & 43.4
    & 38.0
    & -- \\

    \rowcolor{toolbg}
    Mistral-3.2 (24B)
    & Tool
    & 39.0
    & 43.9
    & 44.7
    & 33.8
    & \uline{45.1}
    & \uline{41.4}
    & \textcolor{green!50!black}{+3.4} \\

    \rowcolor{oraclebg}
    & Oracle
    & 82.0
    & 88.3
    & 71.2
    & 33.5
    & \textbf{51.0}
    & \textbf{67.4}
    & \textcolor{green!50!black}{+29.4} \\

    \midrule

    & Raw
    & 32.9
    & 36.7
    & 28.4
    & 38.4
    & 44.8
    & 36.2
    & -- \\

    \rowcolor{toolbg}
    Gemma-3 (12B)
    & Tool
    & --
    & --
    & --
    & --
    & --
    & --
    & -- \\

    \rowcolor{oraclebg}
    & Oracle
    & 86.5
    & 96.7
    & 61.4
    & 34.6
    & \textbf{50.4}
    & \textbf{68.8}
    & \textcolor{green!50!black}{+32.6} \\

    \midrule

    & Raw
    & 30.1
    & 31.9
    & 25.4
    & 30.5
    & 41.0
    & 31.8
    & -- \\

    \rowcolor{toolbg}
    RemoteReasoner
    & Tool
    & --
    & --
    & --
    & --
    & --
    & --
    & -- \\

    \rowcolor{oraclebg}
    & Oracle
    & 73.3
    & 77.8
    & 52.3
    & 28.6
    & \textbf{50.7}
    & \textbf{58.7}
    & \textcolor{green!50!black}{+26.9} \\

    \midrule

    & Raw
    & 24.9
    & 32.2
    & 23.5
    & 33.5
    & 29.5
    & 28.7
    & -- \\

    \rowcolor{toolbg}
    RSThinker
    & Tool
    & --
    & --
    & --
    & --
    & --
    & --
    & -- \\

    \rowcolor{oraclebg}
    & Oracle
    & 46.6
    & 50.6
    & 39.0
    & 29.3
    & \textbf{38.5}
    & \textbf{41.7}
    & \textcolor{green!50!black}{+13.0} \\

    \bottomrule
  \end{tabularx}
\end{table}

\begin{figure}[t]
  \centering
  \includegraphics[width=1\textwidth]{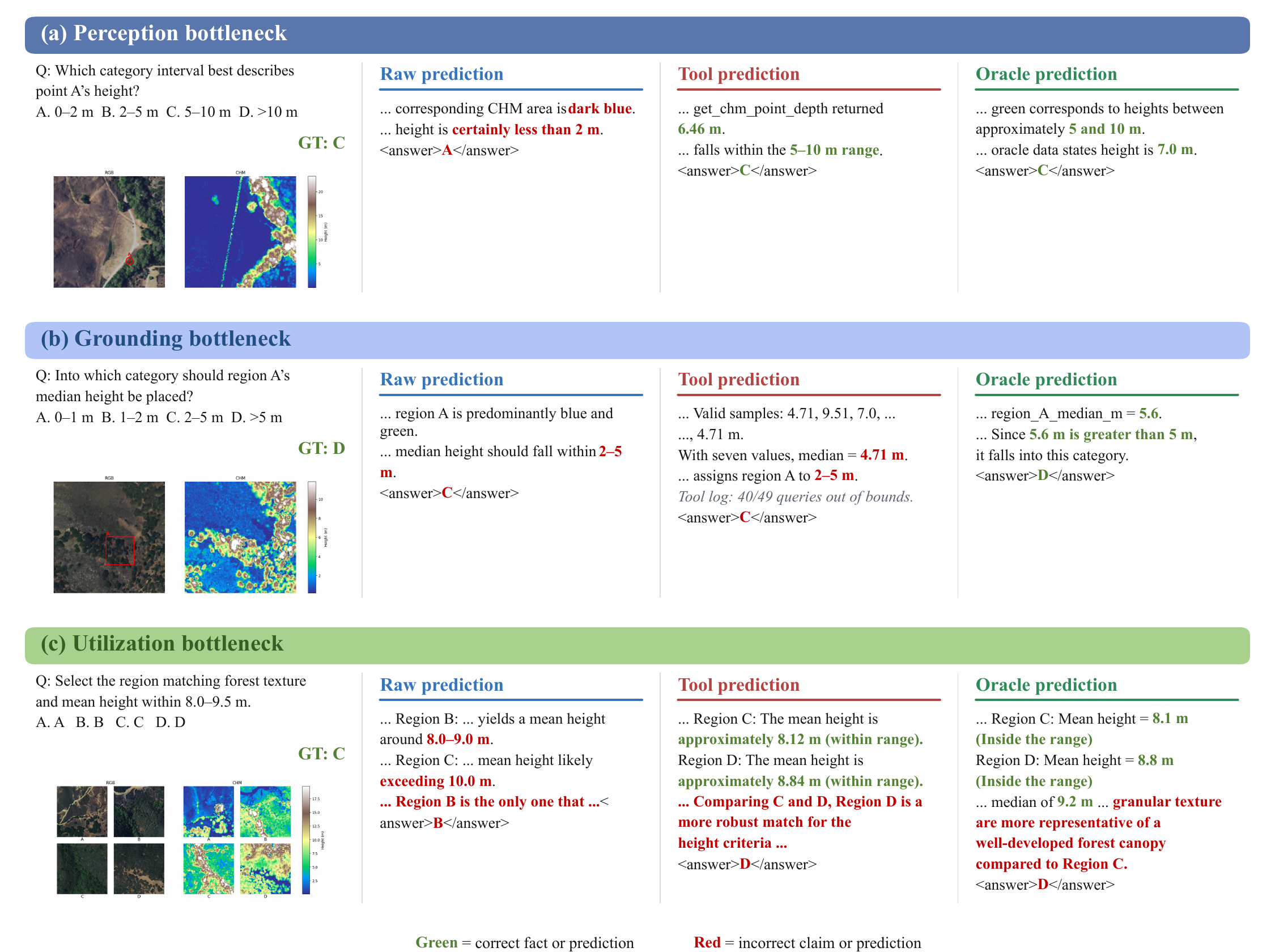}
  \caption{Representative Raw--Tool--Oracle cases.}
  \label{fig:qualitative_rto}
\end{figure}

As shown in Table~\ref{tab:rto_full}, this step-wise intervention reveals three distinct bottlenecks. Fig.~\ref{fig:qualitative_rto} presents representative Raw--Tool--Oracle cases that visualize these step-wise bottlenecks.

The transition from Raw to Tool exposes a Perception bottleneck.
Tool directly returns structured physical measurements and therefore primarily reduces the burden of recovering metric height from rendered CHM representations. Among the seven models with Tool support, six improve in Overall accuracy, with gains ranging from $+2.2$ to $+15.9$ percentage points; L1 also improves for five of the seven models. These results establish that translating CHM color and pixel representations into accurate physical scalars is an important source of noise in the Perception stage (Stage I).

The transition from Tool to Oracle exposes a Grounding bottleneck.
Under Tool, accurate measurements are available, but the model must still determine where to query and associate the returned measurements with the correct spatial target. Oracle further reduces this burden by directly providing geometric facts associated with task-relevant targets or candidates.

Table~\ref{tab:rto_full} reveals an anomalous Tool profile for InternVL3.5-14B, with the performance degradation concentrated on L2 (44.7\% $\rightarrow$ 39.7\%) and L3 (34.1\% $\rightarrow$ 29.6\%), where spatial localization and regional association are more demanding. We therefore further analyze its Tool queries and find a median localization error of 72.1 pixels. The model also frequently adopts coordinates defined over its internal visual resolution rather than the coordinate systems expected by the tools. These observations expose a clear Grounding failure. Even when numerical perception is externalized by Tool, inaccurate spatial localization can prevent the returned physical measurements from being correctly bound to their target regions.

For all seven models with Tool results, Oracle exceeds Tool by 16.0--34.5 percentage points in Overall accuracy.
This recovery pattern indicates that merely exposing numerical height values is insufficient.
The model must also establish a stable correspondence between the measurement and the intended spatial entity.
Spatial Grounding is therefore one of the most critical and fragile bottlenecks in current MLLMs.

The residual error under Oracle exposes a Utilization bottleneck.
Even after Oracle substantially reduces Perception and spatial-association barriers, models remain far from ceiling on downstream tasks. 
While L1 and L2 approach or reach near-perfect performance for several models, Oracle L5 accuracy remains only 38.5\%--78.8\%. 
Thus, even when the relevant physical fact is explicitly available in language and associated with the intended target, models do not consistently convert it into an effective constraint governing the final semantic decision.
The residual Oracle errors therefore expose a distinct failure in the Utilization stage.

\subsection{Counterfactual Height Swap: Testing Evidence Dependence}

\begin{figure}[t]
  \centering
  \IfFileExists{fig7_counterfactual_triplet.pdf}{%
    \includegraphics[width=1\textwidth]{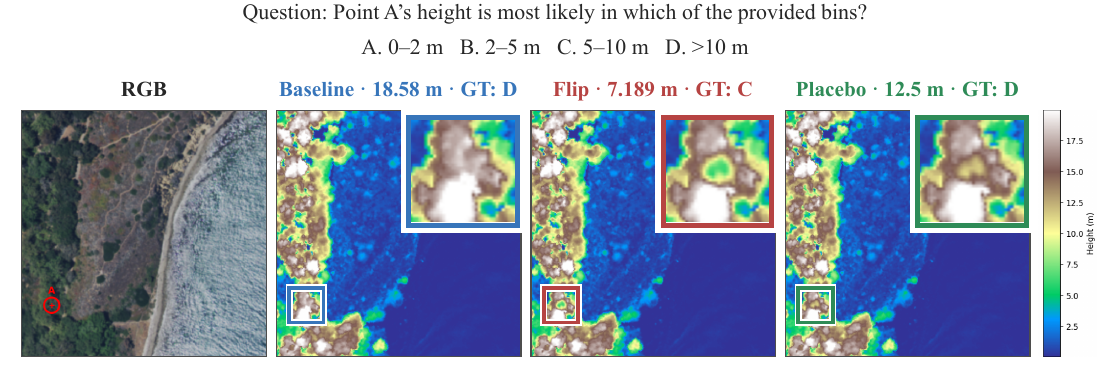}
  }{%
    \fbox{\parbox[c][0.16\textheight][c]{0.94\textwidth}{\centering
    Placeholder for the baseline--flip--placebo counterfactual triplet example.}}
  }
  \caption{Example baseline--flip--placebo triplet. Baseline preserves the original RGB--CHM pair; Flip changes local height across the task decision boundary; Placebo applies a comparable local height edit while preserving the ground truth.}
  \label{fig:counterfactual_triplet_example}
\end{figure}

We further examine whether model decisions establish counterfactual evidence dependence on vertical physical measurements using 133 matched baseline--flip--placebo triplets. As illustrated in Fig.~\ref{fig:counterfactual_triplet_example}, RGB and the valid spatial support are held fixed within each triplet. A Flip locally modifies CHM so that the relevant physical quantity crosses the original decision boundary and changes the ground truth, whereas a Placebo applies the same class of local height perturbation while remaining within the original decision region and therefore preserves the ground truth.

\begin{figure}[t]
  \centering
  \includegraphics[width=\columnwidth]{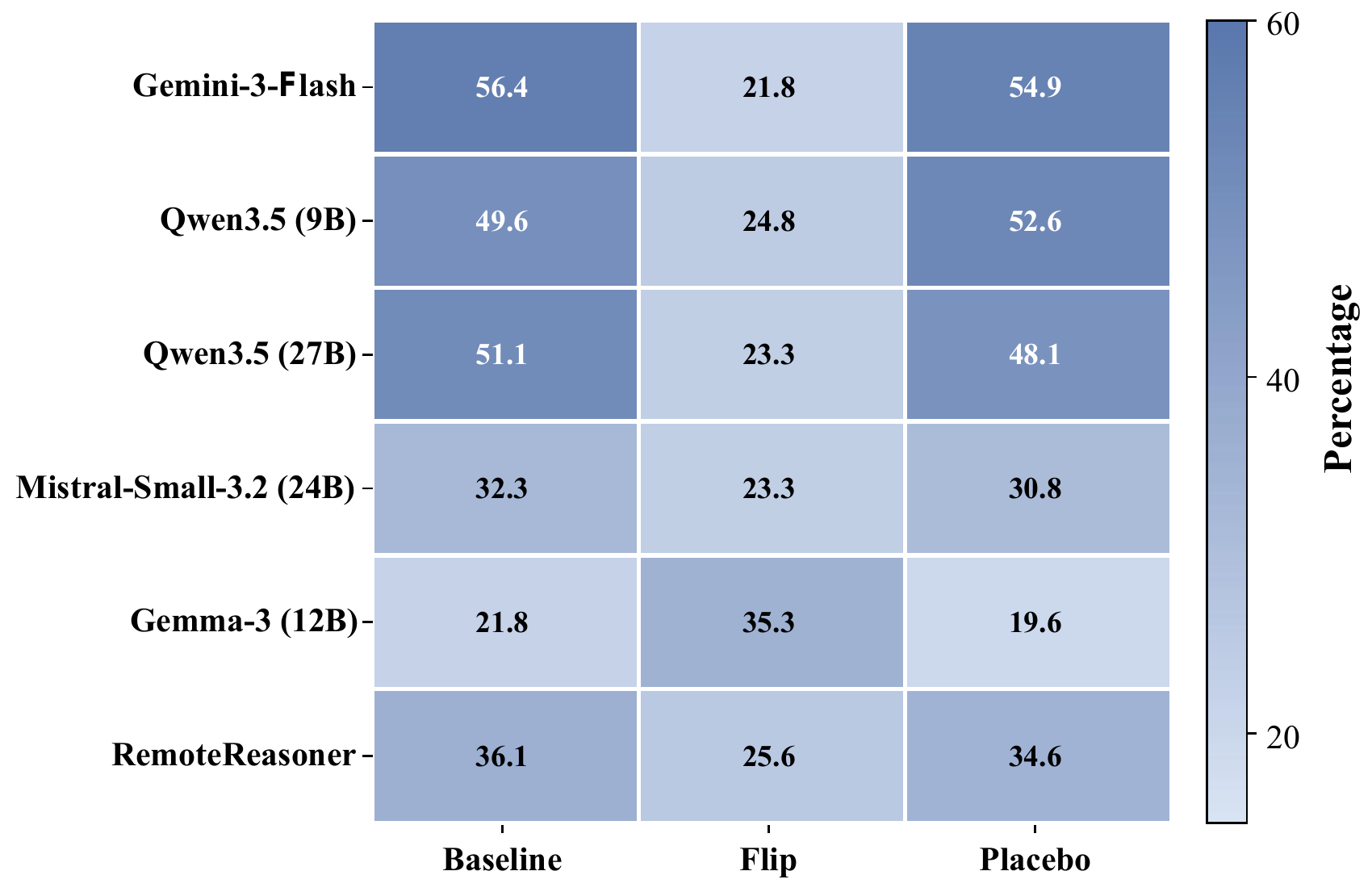}
  \caption{Counterfactual height intervention results.}
  \label{fig:counterfactual_heatmap}
\end{figure}

We report Accuracy under the Baseline, Flip, and Placebo conditions. As shown in Fig.~\ref{fig:counterfactual_heatmap}, five of the six evaluated models remain close to baseline under Placebo but deteriorate sharply under Flip. Gemini-3-Flash drops from 56.4\% Baseline Accuracy to 21.8\% Flip Accuracy, while maintaining 54.9\% Placebo Accuracy; Qwen3.5-9B similarly drops from 49.6\% to 24.8\% Flip Accuracy, while retaining 52.6\% Placebo Accuracy.

This strong asymmetry shows that models often remain stable when a height perturbation should be irrelevant, but fail to update when the altered physical measurement actually changes the correct answer. Current MLLMs therefore do not yet exhibit robust behavioral evidence of dependence on vertical physical evidence and remain susceptible to anchoring on unchanged RGB appearance.

\subsection{Auxiliary Diagnostic: Presentation Robustness of Geometric Evidence}

\begin{figure}[t]
  \centering
  \includegraphics[width=\textwidth]{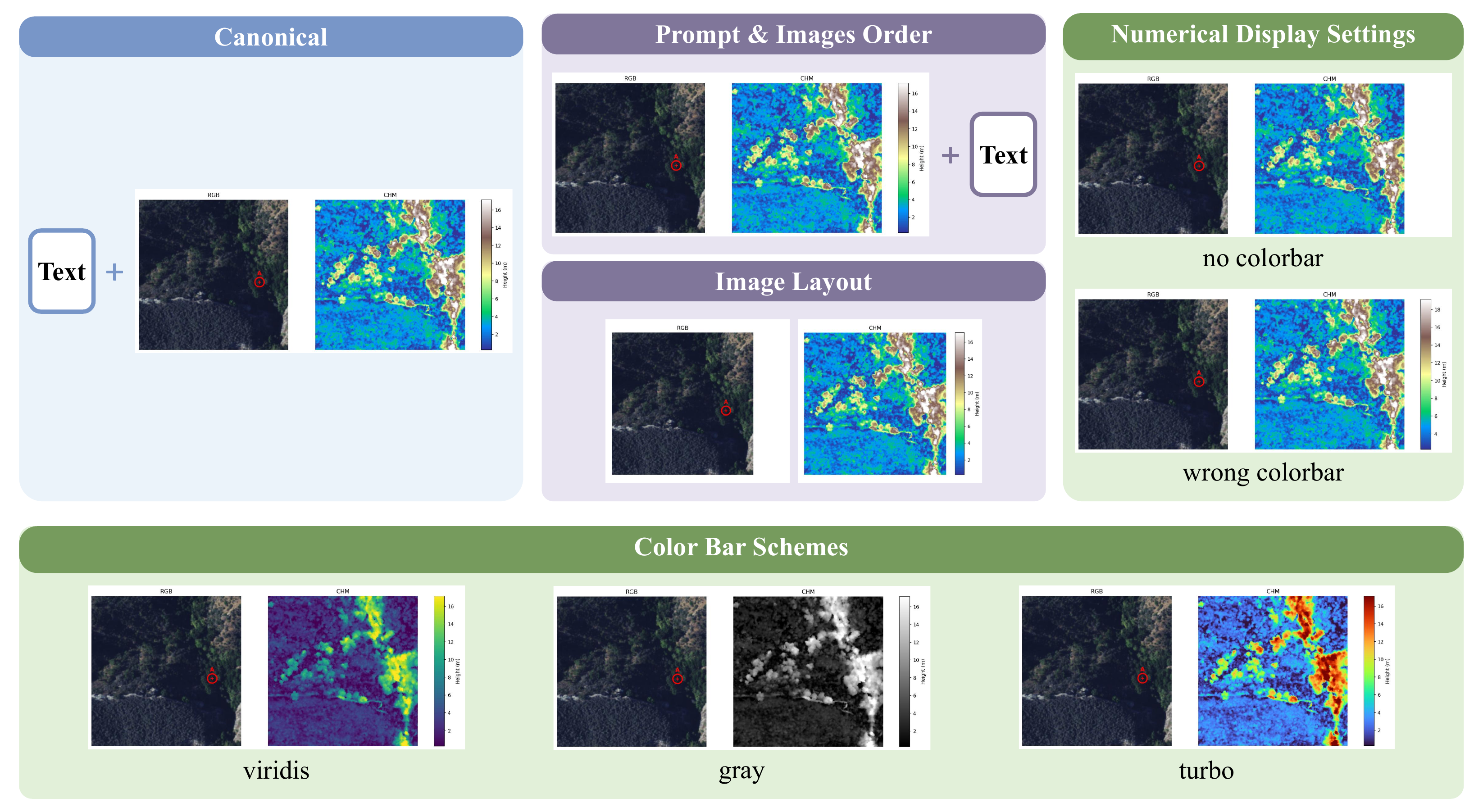}
  \caption{Controlled presentation variants covering Canonical input, Numerical Display Settings, Color Bar Schemes, Prompt \& Images Order, and Image Layout.}
  \label{fig:variants}
\end{figure}

This analysis serves as an auxiliary diagnostic to the central questions above, examining whether part of the observed fragility arises from presentation-level co-registration cues when the underlying evidence is held fixed.
To this end, we test whether models remain stable when the underlying RGB and CHM evidence is unchanged but its presentation is modified. 
The study uses 496 matched instances spanning all 17 tasks. 
As illustrated in Fig.~\ref{fig:variants}, the evaluated conditions cover Canonical input, Numerical Display Settings (no colorbar / wrong colorbar), Color Bar Schemes (viridis / gray / turbo), Prompt \& Images Order, and Image Layout (separate RGB and CHM).

Table~\ref{tab:presentation_robustness} shows that models are substantially more sensitive to spatial layout changes than to colormap or numerical-display variations. Separating RGB and CHM into independent images causes drops of 27.9, 22.2, and 22.9 percentage points for Gemini-3-Flash, Qwen3.5-9B, and Qwen3.5-27B, respectively. In contrast, changing Prompt \& Images Order, switching among viridis, gray, and turbo, or altering colorbar information generally produces smaller model-dependent fluctuations.

The dominant vulnerability lies in maintaining RGB--CHM spatial correspondence across input layouts. Once the physically registered RGB and CHM rasters are separated, cross-image association deteriorates sharply. This indicates that current models still rely heavily on the presentation-level alignment provided by the side-by-side composite, rather than forming a robust geometric co-registration representation that remains stable across presentation formats.

\subsection{Semantic Bias under Semantic--Geometric Conflict}

The previous analyses show that models exhibit emerging capability to access height evidence, together with partial grounding ability.
We now ask the stronger question of which source of evidence governs the final decision when geometric evidence conflicts with a plausible 2D semantic interpretation.
To answer it, we decompose predictions on all 288 L5 instances into Both Correct, Semantic Only, Geometry Only, and Both Wrong.

\begin{table}[t]
  \centering
  \footnotesize
  \caption{Variant robustness under presentation and visual-encoding perturbations. The canonical baseline is \textit{text-first + side-by-side composite + terrain + with colorbar}. Main values show absolute accuracy (\%); superscripts indicate $\Delta$Acc relative to the baseline. \textcolor{green!70!black}{Green} denotes a gain and \textcolor{red!70!black}{red} a drop.}
  \label{tab:presentation_robustness}
  \setlength{\tabcolsep}{3.2pt}
  \begin{tabular*}{\textwidth}{@{\extracolsep{\fill}}lcccccccc@{}}
    \toprule
    & & \multicolumn{2}{c}{\textbf{Presentation Variants}} & \multicolumn{5}{c}{\textbf{Encoding Variants}}\\
    \cmidrule(lr){3-4}\cmidrule(lr){5-9}
    \textbf{Model} & \textbf{Base Acc.} & \textbf{image-first} & \textbf{separate} & \textbf{viridis} & \textbf{gray} & \textbf{turbo} & \textbf{no colorbar} & \textbf{wrong colorbar}\\
    \midrule
    Gemini-3-Flash &56.5&55.7$_{\textcolor{red!80}{-0.8}}$&28.6$_{\textcolor{red!80}{-27.9}}$&57.9$_{\textcolor{green!70!black}{+1.4}}$&54.2$_{\textcolor{red!70}{-2.3}}$&55.0$_{\textcolor{red!70}{-1.5}}$&53.6$_{\textcolor{red!80}{-2.9}}$&53.8$_{\textcolor{red!70}{-2.7}}$\\
    Gemma-3 (12B) &35.7&32.7$_{\textcolor{red!70}{-3.0}}$&31.9$_{\textcolor{red!70}{-3.8}}$&32.3$_{\textcolor{red!70}{-3.4}}$&34.5$_{\textcolor{red!60}{-1.2}}$&34.3$_{\textcolor{red!60}{-1.4}}$&32.5$_{\textcolor{red!70}{-3.2}}$&33.3$_{\textcolor{red!70}{-2.4}}$\\
    Mistral-Small-3.2 (24B) &35.3&34.3$_{\textcolor{red!70}{-1.0}}$&31.9$_{\textcolor{red!70}{-3.4}}$&36.1$_{\textcolor{green!70!black}{+0.8}}$&36.5$_{\textcolor{green!60!black}{+1.2}}$&34.9$_{\textcolor{red!70}{-0.4}}$&35.1$_{\textcolor{red!70}{-0.2}}$&35.3$_{+0.0}$\\
    Qwen3.5 (9B) &50.8&51.8$_{\textcolor{green!60!black}{+1.0}}$&28.6$_{\textcolor{red!80}{-22.2}}$&50.0$_{\textcolor{red!60}{-0.8}}$&50.2$_{\textcolor{red!60}{-0.6}}$&49.8$_{\textcolor{red!60}{-1.0}}$&44.6$_{\textcolor{red!80}{-6.2}}$&50.4$_{\textcolor{red!50}{-0.4}}$\\
    Qwen3.5 (27B) &53.8&52.3$_{\textcolor{red!60}{-1.5}}$&30.9$_{\textcolor{red!80}{-22.9}}$&51.3$_{\textcolor{red!70}{-2.5}}$&52.8$_{\textcolor{red!60}{-1.0}}$&51.6$_{\textcolor{red!70}{-2.2}}$&48.9$_{\textcolor{red!75}{-4.9}}$&49.4$_{\textcolor{red!70}{-4.4}}$\\
    RemoteReasoner &29.8&32.7$_{\textcolor{green!70!black}{+2.9}}$&--&28.6$_{\textcolor{red!60}{-1.2}}$&32.5$_{\textcolor{green!65!black}{+2.7}}$&29.8$_{+0.0}$&28.8$_{\textcolor{red!60}{-1.0}}$&29.0$_{\textcolor{red!50}{-0.8}}$\\
    \bottomrule
  \end{tabular*}
\end{table}

\begin{figure}[t]
  \centering
  \includegraphics[width=\columnwidth]{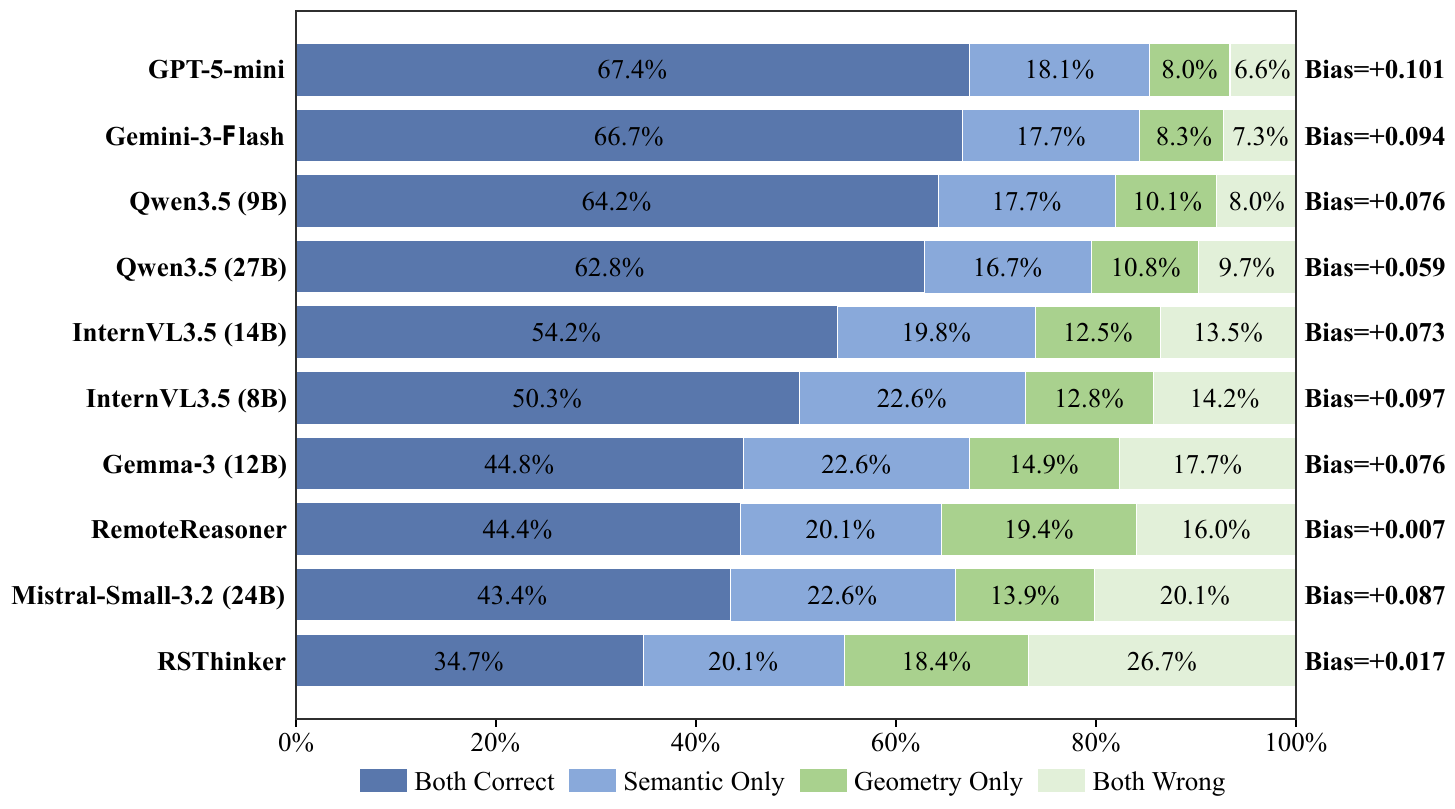}
  \caption{Semantic--geometric error decomposition.}
  \label{fig:l5_error_decomposition}
\end{figure}

As shown in Fig.~\ref{fig:l5_error_decomposition}, Semantic Only exceeds Geometry Only for all evaluated models. The gap reaches 10.1 points for GPT-5-mini, 9.4 points for Gemini-3-Flash, and 7.6 points for Qwen3.5-9B. RemoteReasoner is nearly balanced, with a difference of only 0.7 points, although its Both Correct rate remains below that of the strongest general-purpose models.

These results reveal a pronounced Semantic Appearance Bias. When semantic appearance and geometric height cannot jointly support the same decision, most models are more likely to preserve the interpretation suggested by 2D appearance than to obey the acquired physical height constraint. Height facts may therefore be accessible without consistently acquiring sufficient decision authority to override 2D semantic priors.

\subsection{Defining the Vertical Structure Utilization Gap}

Taken together, the preceding diagnostics identify a continuous breakdown in the flow of vertical physical evidence, spanning physical-value access from the raw representation, spatial grounding to the correct visual entity, and downstream utilization of grounded evidence in the final decision.
Current MLLMs can already acquire part of the underlying geometric information, yet that evidence fails to remain reliable as it proceeds through raw perception, spatial co-registration, conversion into geometric logical constraints, and ambiguity-resolving semantic decision-making. Perception remains affected by visual measurement noise, Grounding is highly fragile, and even after these upstream barriers are substantially reduced, Utilization still fails to consistently turn explicit physical facts into hard constraints on the final decision.

We formally define this systematic breakdown as the Vertical Structure Utilization Gap, namely the failure of available and accessible vertical physical evidence to be reliably propagated through Perception, Grounding, and Utilization until it becomes an effective constraint governing the final ambiguity-resolving semantic decision.
The Vertical Structure Utilization Gap is therefore not a single-stage error, but a progressive loss of physical evidence authority along the representation-to-decision pathway.

\section{Towards Height-Aware MLLMs}
\label{sec:future}

The preceding diagnostics suggest that the Vertical Structure Utilization Gap does not arise from a single task-specific weakness. Instead, it exposes two more fundamental architectural bottlenecks. Models must first recover metrically meaningful physical evidence from the supplied height representation, and the acquired measurements must then be spatially co-registered with the correct 2D semantic entities before they can participate in Utilization. These findings motivate reconsidering how future height-aware MLLMs represent, align, and integrate physical structure.

\subsection{The Representation Bottleneck: Pixel-to-Physics Transformation}

Under the Raw Visual pathway adopted in this work, CHM is exposed to the model as a color-mapped visualization with a metric colorbar, requiring physical height measurements to be recovered from a pixel-based representation. As shown by the presentation-robustness results in Table~\ref{tab:presentation_robustness}, the canonical Raw Visual setting achieves the highest cross-model average performance among the common presentation and encoding variants, and therefore serves as a standardized and comparatively favorable passive visual reference. Against this reference, the Raw$\rightarrow$Tool recovery in Table~\ref{tab:rto_full} further indicates that, even under the strongest visual presentation considered, transforming visual representations into metric physical quantities remains a non-trivial perceptual bottleneck.

One possible explanation lies in the objectives of generic vision--language representation learning.
CLIP, for example, learns transferable visual representations through natural-language supervision and image--text correspondence~\cite{radford2021clip}. Such objectives primarily target abstract 2D visual semantics and semantic alignment rather than the explicit recovery of absolute physical scale or depth scalars. 
A more cautious interpretation is therefore that absolute scale and calibrated metric relations may be weakened during semantic abstraction rather than being explicitly preserved by the training objective.
We view this as a hypothesis rather than a direct causal explanation of the observed failures.

This motivates an explicit Pixel-to-Physics Transformation. A height-aware MLLM would need to move beyond relying solely on recovering physical scalars indirectly from rendered colors and pixel patterns. Instead, geometric representations should preserve the measurement value, physical unit, spatial support, and coordinate reference, allowing height to enter the model as an explicitly physical quantity before subsequent multimodal reasoning.

\subsection{The Grounding Bottleneck: Latent Spatial Co-registration}

The diagnostics in Section~\ref{sec:diagnostics} further indicate that, once physical height values become accessible, the dominant difficulty shifts toward spatial binding between physical measurements and visual entities. 
This conclusion is jointly supported by the Tool$\rightarrow$Oracle separation in Table~\ref{tab:rto_full} and the layout sensitivity in Table~\ref{tab:presentation_robustness}. The former shows that structured numerical access alone does not resolve target localization, while the latter shows that data-level RGB--CHM registration does not automatically yield stable cross-modal correspondence when the presentation layout changes.

Future MLLMs may therefore benefit from Latent Spatial Co-registration rather than relying primarily on text-level coordinate estimation and external API calls. RGB semantic features and height-geometric features should share an explicit spatial reference within the latent representation, supporting bidirectional correspondence between visual entities and physical measurements while preserving that alignment throughout multimodal fusion. Shared coordinate embeddings, tied spatial grids, coordinate-preserving cross-attention, and common spatial indices are possible realizations. The central objective is to make spatial co-registration an architectural property of the representation rather than an implicit consequence of input layout.

\subsection{Future Architectural Directions}

The two bottlenecks above suggest three potential architectural directions toward physically grounded, height-aware MLLMs.

For unified physical--semantic tokenization, future tokenizers could jointly encode continuous geometric scalars, semantic information, and spatial locations into a unified token sequence. Rather than representing height only through RGB-like visual pixels, a physical token could simultaneously carry its actual measurement, corresponding spatial location, and associated visual entity. Such a unified representation would shorten the transformation path from physical measurement to semantic reasoning and reduce information loss introduced by appearance-based rendering.

For multi-modal geometry encoding, a dedicated geometry encoder could be introduced alongside the conventional visual backbone to preserve the native structure of physical observations. Depending on the sensing modality, this branch could process continuous height rasters, depth fields, LiDAR point clouds, or other geometric measurements, and subsequently fuse metric-aware geometric features with visual semantic tokens. Decoupling geometric representation learning from appearance-oriented 2D visual encoding could provide more stable physical-structure information for downstream reasoning.

For geometry-grounded contrastive alignment, cross-modal alignment objectives could further incorporate explicit geometric consistency constraints. RGB--geometry regions with consistent spatial and physical relationships should be pulled closer in representation space, whereas regions with similar visual semantics but inconsistent geometric relationships could serve as hard negatives. Counterfactual height perturbations could further constrain model behavior. Predictions should change when the underlying physical condition changes, while remaining invariant to decision-irrelevant perturbations. In this way, cross-modal learning could move beyond generic semantic-similarity matching toward cross-modal alignment explicitly constrained by physical geometry.

Overall, these directions suggest that future MLLMs should move beyond an appearance-centric modeling paradigm toward architectures that can natively represent physical measurements, perform latent spatial co-registration, and explicitly integrate geometric facts into semantic reasoning. A genuinely height-aware MLLM would therefore need to do more than simply accept an additional CHM image; physical geometric evidence would need to become a first-class information source throughout the representation-to-decision pipeline.

\section{Conclusion}

Reliable remote-sensing natural-scene understanding requires moving beyond 2D optical appearance. To systematically investigate whether MLLMs can effectively exploit vertical physical evidence under appearance ambiguity, we introduced VertiCue-Bench, the first diagnostic benchmark for remote-sensing natural-scene understanding to use controlled interventions, together with a three-stage Perception--Grounding--Utilization framework that traces the complete evidence pathway from physical measurement acquisition and spatial co-registration to semantic decision-making. Through controlled diagnostic interventions, we formally define and substantiate the Vertical Structure Utilization Gap. Although current MLLMs exhibit emerging capability to access vertical geometric information, physical evidence remains systematically degraded as it proceeds through perception, spatial grounding, and geometry-constrained semantic decision-making, manifesting as impaired physical-value acquisition, fragile spatial co-registration, and ineffective conversion of geometric facts into decision constraints. These findings suggest that the next generation of height-aware MLLMs may benefit from moving beyond appearance-centric modeling toward architectures in which physical measurements are natively represented, stably grounded, and explicitly integrated into high-level semantic reasoning, providing a systematic scientific and diagnostic foundation for physically grounded and geometry-aware remote-sensing MLLMs.
The central challenge is therefore not simply to provide additional geometric modalities, but to ensure that physical measurements remain spatially grounded and decision-relevant throughout the reasoning process.

\label{ReferencesStart}

\end{document}